\documentclass[10pt,twocolumn,letterpaper]{article}

\usepackage[pagenumbers]{cvpr} 

\usepackage[dvipsnames]{xcolor}

\definecolor{cvprblue}{rgb}{0.21,0.49,0.74}
\usepackage[pagebackref,breaklinks,colorlinks,citecolor=cvprblue]{hyperref}

\usepackage{acronym}
\usepackage{wrapfig}
\usepackage{multirow}
\usepackage[export]{adjustbox}
\usepackage{rotating}
\usepackage{mathtools}
\usepackage{tabularx}
\usepackage{enumitem}

\def\paperID{*****} 
\def\confName{3DV\xspace}
\def\confYear{2026\xspace}

\title{HDR Environment Map Estimation with Latent Diffusion Models}

\author{Jack Hilliard\\
{\tt\small jh00695@surrey.ac.uk}
\and
Adrian Hilton\\
{\tt\small a.hilton@surrey.ac.uk}\\
University of Surrey\\
\and
Jean-Yves Guillemaut\\
{\tt\small j.guillemaut@surrey.ac.uk}\\
}

\begin{document}
\newacro{bmvc}[BMVC]{British Machine Vision Conference}
\newacro{iccv}[ICCV]{International Conference on Computer Vision}

\newacro{ai}[AI]{Artificial Intelligence}
\newacro{ar}[AR]{Augmented Reality}
\newacro{sdk}[SDK]{Software Development Kit}
\newacro{fps}[FPS]{Frames Per Second}
\newacro{nlp}[NLP]{Natural Language Processing}
\newacro{lfov}[LFOV]{Limited Field of View}
\newacro{fov}[FOV]{Field of View}
\newacro{gan}[GAN]{Generative Adversarial Network}
\newacro{vit}[ViT]{Vision Transformer}
\newacro{mlp}[MLP]{Multilayer Perceptron}
\newacro{ibl}[IBL]{Image-Based Lighting}
\newacro{hdr}[HDR]{High Dynamic Range}
\newacro{hdri}[HDRI]{High Dynamic Range Image}
\newacro{ldr}[LDR]{Low Dynamic Range}
\newacro{ldri}[LDRI]{Low Dynamic Range Image}
\newacro{erp}[ERP]{Equirectangular Panorama}
\newacro{ldm}[LDM]{Latent Diffusion Model}
\newacro{cnn}[CNN]{Convolutional Neural Network}
\newacro{lstm}[LSTM]{Long Short Term Memory}
\newacro{tsp}[TSP]{Temporal Spatial Predictor}
\newacro{dit}[DiT]{Diffusion Transformer}
\newacro{mmdit}[MMDiT]{Multi-Modal Diffusion Transformer}
\newacro{rct}[RCT]{Recurrent Content Transfer}
\newacro{sienet}[SIENET]{Siamese Image Expansion Network}
\newacro{brdf}[BRDF]{Bidirectional Reflectance Distribution Function}
\newacro{sg}[SG]{Spherical Gaussian}
\newacro{sh}[SH]{Spherical Harmonic}
\newacro{mr}[MR]{Mixed Reality}
\newacro{er}[ER]{Extended Reality}
\newacro{ar}[AR]{Augemented Reality}
\newacro{vr}[VR]{Virtual Reality}
\newacro{fid}[FID]{Fréchet Inception Distance}
\newacro{rmse}[RMSE]{Root Mean Square Error}
\newacro{sirmse}[si-RMSE]{scale invariant Root Mean Square Error}
\newacro{psnr}[PSNR]{Peak Signal-to-Noise Ratio}
\newacro{wmsa}[W-MSA]{Window multi-head self-attention layer}
\newacro{pswmsa}[PSW-MSA]{Panoramically warped Shifted Window Multi-head Self-Attention layer}
\newacro{pam}[PAM]{Pitch attention Module}
\newacro{vae}[VAE]{Variational Autoencoder}
\newacro{cae}[CAE]{Contractive Autoencoder}
\newacro{rbm}[RBM]{Restricted Boltzmann Machine}
\newacro{gcn}[GCN]{Graph Convolutional Network}
\newacro{svm}[SVM]{Support Vector Machine}
\newacro{srbf}[SRBF]{Spherical Radial Basis Function}
\newacro{nerf}[NERF]{Neural Radiance Fields}
\newacro{std}[STD]{Spherical Transport Distance}
\newacro{ddpm}[DDPM]{Denoising Diffusion Probabilistic Model}
\newacro{lae}[LAE]{Latent Autoencoder}

\maketitle
\begin{abstract}
We advance the field of HDR environment map estimation from a single-view image by establishing a novel approach leveraging the Latent Diffusion Model (LDM) to produce high-quality environment maps that can plausibly light mirror-reflective surfaces. A common issue when using the ERP representation, the format used by the vast majority of approaches, is distortions at the poles and a seam at the sides of the environment map. We remove the border seam artefact by proposing an ERP convolutional padding in the latent autoencoder. Additionally, we investigate whether adapting the diffusion network architecture to the ERP format can improve the quality and accuracy of the estimated environment map by proposing a panoramically-adapted Diffusion Transformer architecture. Our proposed PanoDiT network reduces ERP distortions and artefacts, but at the cost of image quality and plausibility. We evaluate with standard benchmarks to demonstrate that our models estimate high-quality environment maps that perform competitively with state-of-the-art approaches in both image quality and lighting accuracy.
\end{abstract}    
\section{Introduction}
\label{sec:intro}
The research on illumination estimation from a \ac{lfov} \ac{ldri} for \ac{ibl} has followed the trends in image generation to improve the quality of the estimated environment maps. Image quality is crucial when using environment maps to light mirror reflective surfaces, as the details in the environment map will be visible on the rendered surface. 
Consequently, any image artefacts or implausibility in the estimated environment map will reduce the plausibility of the composited object. 
Trends in image generation \cite{Esser2021, Zhao2021comodgan, dosovitskiy2021, Wang2022UFormer, Karras2020CVPR} have often been adopted in the field of illumination estimation to improve environment map generation \cite{Wang2022stylelight, Akimoto2022, Dastjerdi2022, Dastjerdi2023}. Recent image generation models have used \ac{ldm}s to great effect; Latent Diffusion, proposed by Rombach \etal \cite{Rombach2022}, and Stable Diffusion \cite{Podell2024sdxl, Sauer2024, Esser2024} can both produce images nearly indistinguishable from those captured with a camera or created by an artist. 
These models have consistently proven their adaptability to different tasks, including the field of inpainting and outpainting 
\cite{Li2024image, Lugmayr2022, Podell2024sdxl, Rombach2022, Sauer2024, Wu2024, Xie2023}.

\begin{figure}[t]
    \centering
        \begin{subfigure}[b]{0.9\linewidth}
		\centering
        \centerline{\includegraphics[width=\textwidth,keepaspectratio]{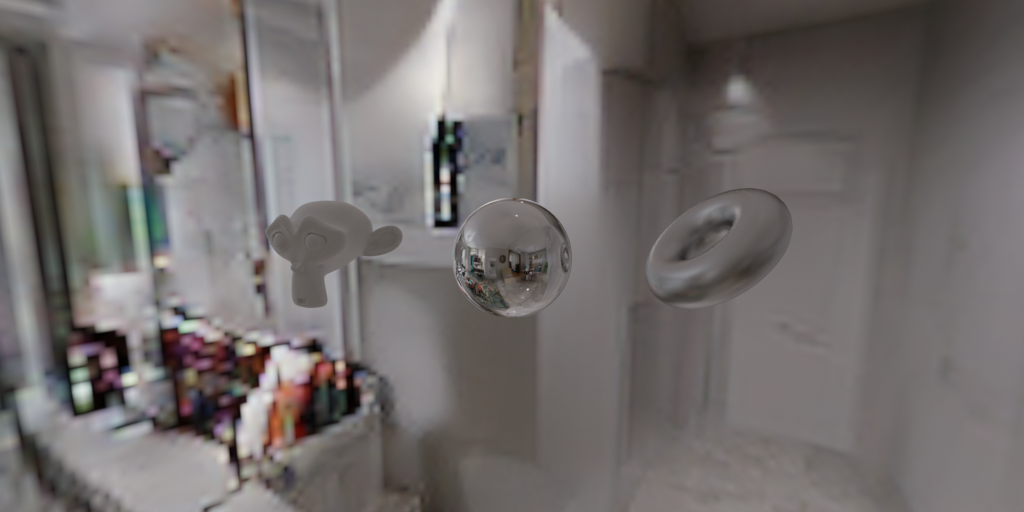}}
    \end{subfigure}
    \hfill
    \begin{subfigure}[b]{0.9\linewidth}
	\centering
        \centerline{\includegraphics[width=\textwidth,keepaspectratio]{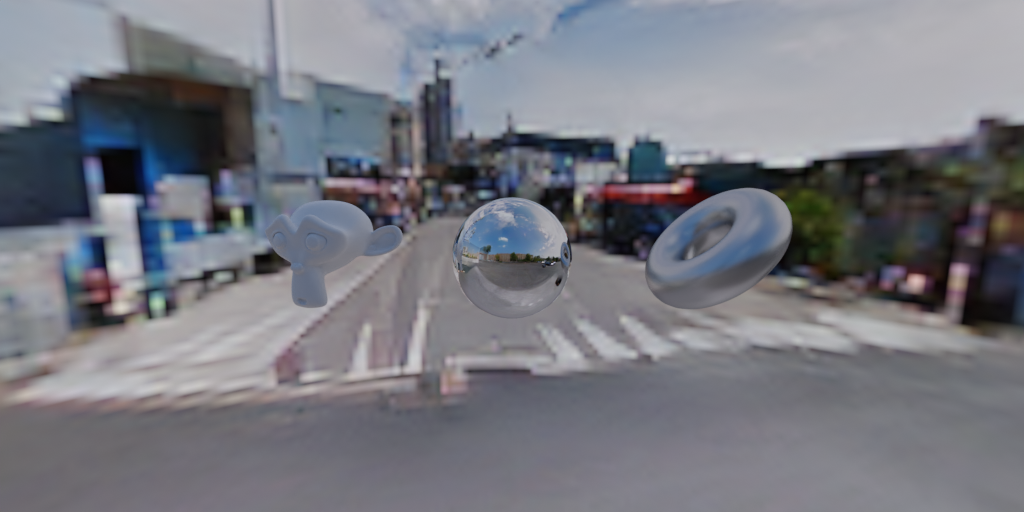}}
    \end{subfigure}
\caption{Examples of objects with different surface properties rendered with indoor (\textit{top}) and outdoor (\textit{bottom}) \acs{hdr} environment maps generated from our LDM with U-Net backbone.}
\label{fig:object_render}
\end{figure}

The standard image format for \ac{ibl} illumination estimation is the 360$^\circ$ \ac{erp} image as it contains the illumination conditions for the whole scene at the camera position in a single image \cite{Akimoto2022, Chen2024, Dastjerdi2023, Gardner2017, Gardner2019, Somanath2021, Song_2019_CVPR, Weber2022, Zhang2017ICCV}. Various approaches have been proposed to adapt neural networks to the \ac{erp} image format and can be categorised as image manipulations at inference, loss function tailoring or architecture adaptations. A few models have been proposed for applying \ac{ldm}s to \ac{erp} image generation \cite{Chen2024, Lu2024, Wang2024, Wu2024}, however, they have not adapted the \ac{ddpm} architecture to \ac{erp}s as they use foundational models that prevent this. Consequently, \ac{erp} artefacts have not been removed or inference time and memory requirements have increased.

We present an \ac{ldm} that estimates the \ac{hdr} environment map, which can be used for \ac{ibl}, from an \ac{lfov} \ac{ldri} and its associated \ac{lfov} mask in \ac{erp} format. It works at a resolution of 256 by 512 for both indoor and outdoor environments. For the \ac{lae}, we propose an \ac{erp} convolutional padding that removes the border seam introduced when encoding and decoding an \ac{erp} image. Additionally, we propose two variations of the model: one with the \ac{ddpm} U-Net of \cite{Rombach2022}, the other, a panoramically adapted \ac{dit}, PanoDiT, that uses a Diffusion Pitch Attention Module (DPAM) that employs the PAM from PanoSWIN \cite{Ling2023} to improve the network's understanding of the \ac{erp} representation.

We evaluate each model's ability to remove \ac{erp} artefacts by observing rotated panoramas and cube-map representations to compare the common locations where \ac{erp} artefacts and warping occur. We quantitatively compare using standard illumination estimation benchmarks proposed by Weber \etal \cite{Weber2022}. Through this evaluation, we demonstrate that both our models perform competitively with state-of-the-art approaches. We also note that the \ac{ddpm} with U-Net architecture produces environment maps with better image quality and the PanoDiT architecture estimates lighting more accurately. Furthermore, we compare against current state-of-the-art illumination estimation methods to demonstrate the ability of our models. Examples of rendered objects with our environment maps are presented in \Cref{fig:object_render}.
Our contributions are summarised as follows:
\begin{itemize}[topsep=5pt]
    \setlength{\itemsep}{5pt}
    \item To the author's knowledge, this work is the first implementation of an \ac{ldm} to the task of \ac{hdr} environment map estimation from an \ac{lfov} \ac{ldri} in \ac{erp} image format.
    \item We present a novel convolutional padding and \ac{ddpm} architecture designed for the \ac{erp} image format. These remove the border seam and homogenise generation at the sides of the image, creating a continuous panorama.
    \item We evaluate our models with standard benchmarks to demonstrate that they perform at a state-of-the-art level, reducing the \ac{erp} artefacts and accurately rendering diffuse, glossy and mirror reflective surfaces.
\end{itemize}

\section{Related Work}\label{LDM:lit_review}
\noindent\textbf{Illumination Estimation}\hspace{15pt}The field of illumination estimation from a single-view image is a long-standing field of computer vision and neural rendering. This field aims to estimate the full 360$^\circ$ lighting conditions of a scene at the point of the camera's location from a single \ac{lfov} \ac{ldri}. Classically, this was achieved by photographing multiple light probes of different surface reflectance properties from different angles and stitching the captured images into a 360$^\circ$ \ac{erp} environment map \cite{Debevec1998}. However, this approach is not viable for real-time rendering applications such as \ac{mr} object compositing. The seminal work of Gardner \etal~\cite{Gardner2017} first proposed the use of a \ac{cnn} autoencoder to estimate the \ac{hdr} environment map of a scene from a single \ac{ldri}. Recent illumination estimation works have opted for an \ac{ibl} approach \cite{Gardner2017, Zhang2017ICCV, Song_2019_CVPR, Somanath2021, Hilliard2023, Hilliard2024, Dastjerdi2022, Dastjerdi2023, Akimoto2022, Wang2022stylelight, Weber2022} over a basis function regression \cite{Gardner2019, Hold-Geoffroy2017, Hold-Geoffroy2019, Zhang2019LightEst, Cheng2018, Garon2019, Gkitsas2020, Li2019, Bai2022, zhan2021emlight, Zhan2021ICCV, Zhan2022} as the \ac{hdr} environment maps used in \ac{ibl} contain the detail required to render mirror reflective objects with features from the scene appearing on their surface. To further enhance the plausibility of the environment map for rendering mirror-reflective objects, there has been a focus on improving image quality by utilising state-of-the-art models from the field of image generation. StyleLight \cite{Wang2022stylelight} used a modified StyleGANV2 \cite{Karras2020CVPR}, Akimoto \etal \cite{Akimoto2022} implemented a VQGAN \cite{Esser2021}, Hilliard \etal \cite{Hilliard2024} proposed a U-Net style \ac{vit} \cite{dosovitskiy2021, Wang2022UFormer} and EverLight \cite{Dastjerdi2023} and ImmerseGAN \cite{Dastjerdi2022} a modified version of the CoModGAN \cite{Zhao2021comodgan}.

We leverage the current state-of-the-art in image generation architecture, the \ac{ldm}, to further advance the field of illumination estimation and improve the image quality using a smaller training dataset than the current state-of-the-art.

\noindent\textbf{Latent Diffusion Models}\hspace{15pt}
\ac{ldm}s \cite{Rombach2022} have proven to produce realistic images with a large variation in their semantic distribution and have since been adapted and applied to various computer vision applications including image inpainting and outpainting by \cite{Li2024image, Lugmayr2022, Podell2024sdxl, Rombach2022, Sauer2024, Wu2024, Xie2023}. 
Stable Diffusion models \cite{Podell2024sdxl, Sauer2024} use the mask and encoded masked image as the condition of the denoising process through a hybrid method of concatenation at the beginning of each denoising step and cross-attention throughout the \ac{ddpm}. \ac{ldm}s have also been applied to generate and manipulate 360$^\circ$ \ac{erp}s \cite{Lu2024,Wang2024,Wu2024}. StitchDiffusion \cite{Wang2024} adopts the MultiDiffusion model to customise 360$^\circ$ panoramas with text prompts. AOG-Net \cite{Lu2024} outpaints 360$^\circ$ \ac{erp}s from an \ac{lfov} image by progressively outpainting a local region in an autoregressive manner. Each local region is outpainted with an \ac{ldm} backbone \cite{Rombach2022} using a text prompt, global cubemap, spherical geometry cubemap projection and the \ac{lfov} to be outpainted. PanoDiffusion \cite{Wu2024} builds on \cite{Rombach2022} to outpaint both the RGB and depth map in 360$^\circ$ \ac{erp} from an \ac{lfov} RGB input image. Although both \cite{Wang2024} and \cite{Wu2024} design their methods for \ac{erp} images, they do not completely remove the border seam artefact.

\begin{figure*}[t]
    \centering
    \begin{subfigure}[b]{0.35\linewidth}
        \centering
        \centerline{\includegraphics[width=\textwidth]{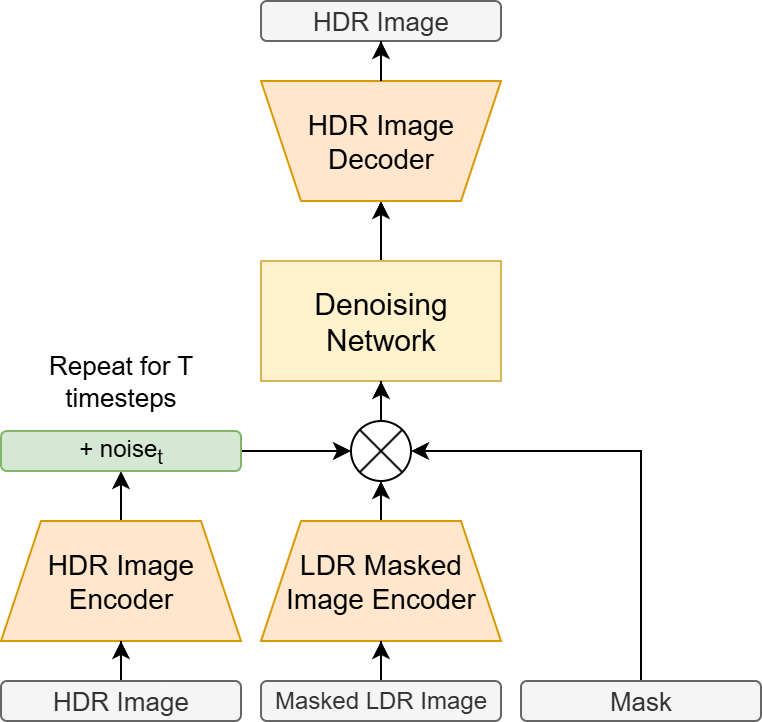}}
        \caption{Training}\medskip
        \label{fig:LDM_train}
    \end{subfigure}
    \begin{subfigure}[b]{0.35\linewidth}
        \centering
        \centerline{\includegraphics[width=\textwidth]{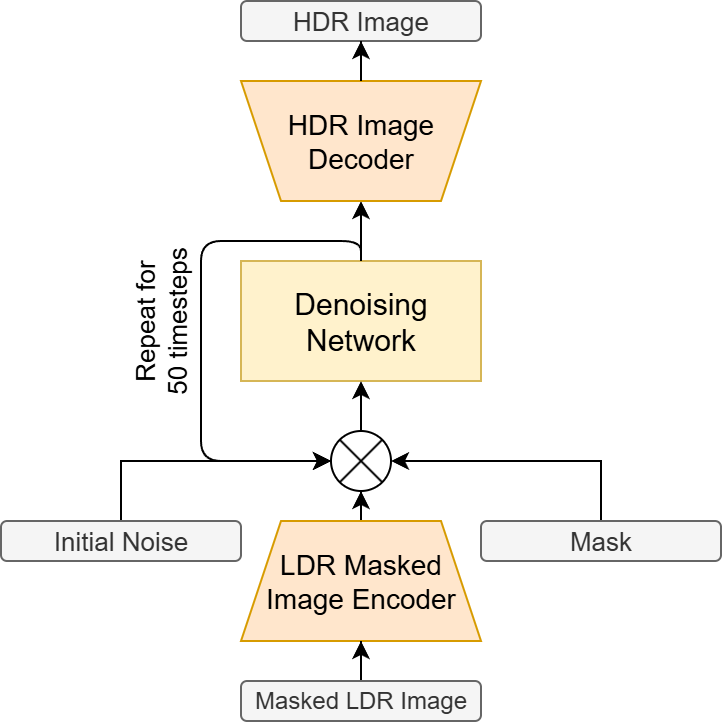}}
        \caption{Inference}\medskip
        \label{fig:LDM_infer}
    \end{subfigure}
\caption{Summary of the training and inference processes. During training, the model applies noise to the latent HDR Image at each timestep. During inference, random noise is used as the initial input to the model and concatenated with the conditions. At each successive timestep, the output of the Denoising Network is used in place of the initial noise.}
\label{fig:model}
\end{figure*}

\noindent\textbf{Adapting Deep Learning Models to 360\textdegree~ERP}\hspace{15pt}360$^\circ$ \ac{erp}s are spherical surfaces warped to be presented on a 2D plane and, as such, represent the illumination conditions at a point in a scene in image format. However, this format incurs warped areas at the top and bottom of the image and includes the consideration that the sides of the image are connected. When using this format in neural networks, artefacts can often appear where objects are not correctly generated at the poles of the image and the sides are not homogenously generated, often with a seam at the border. Solutions to remove these artefacts generated by deep learning models can be put into three categories: image manipulations \cite{Akimoto2019, Akimoto2022, Chou2020, Dastjerdi2023, Feng2022, Gond2023, Hara2021, Hilliard2023, Kim2021, Li2024image, Somanath2021, Wang2024, Wu2024, Yun2022}, tailoring loss functions \cite{Akimoto2022, Dastjerdi2022, Dastjerdi2023, Somanath2021} and adapting architectures \cite{Su2017, Cohen2018, Coors2018, Hilliard2024, Su2019, Shen2022, zhan2021emlight}.

Adapting the neural network architecture has proven advantageous as it removes the need for the network to learn the warped formatting. These adaptations can remove border seams and warping at the poles with the added benefits of higher-quality image generation. Initial approaches adapted the kernel shape of convolutional neural networks \cite{Su2017, Cohen2018, Coors2018, Su2019, Su2022}.
Later adaptations were applied to the \ac{vit} architecture. PanoFormer \cite{Shen2022} adapted a \ac{vit} attention block with a panoramic structure-guided transformer block. Attention was performed between a central token and 8 surrounding tokens to prevent warping. This adopted a UFormer \cite{Wang2022UFormer} style architecture with circular padding for horizontal sides during convolutions. Ling \etal \cite{Ling2023} observed that attention windows at the poles of the \ac{erp} should be considered connected. They proposed PanoSWIN, which changed the shifted window attention layout to account for the connections at the \ac{erp} borders and implemented a pitch attention module to further account for the \ac{erp} warping at the poles.

Currently, in the field of illumination estimation, only EMLight~\cite{zhan2021emlight} and 360U-Former~\cite{Hilliard2024} have adapted the model architecture to the \ac{erp} representation. We compare the use of the pitch attention module in a \ac{dit} against a standard U-Net to compare how adapting the \ac{ddpm} architecture affects the quality of the generated environment map. Furthermore, we explore the use of adapted convolutional layers in the \ac{lae} to remove \ac{erp} border artefacts, which are present in the majority of illumination estimation approaches.

\section{Methodology}\label{LDM:methodology}

We propose a novel method to estimate the HDR environment map of a scene, at the camera position, from an \ac{lfov} \ac{ldri} building on the \ac{ldm} design of Rombach \etal \cite{Rombach2022} and masked outpainting approach of \cite{Podell2024sdxl, Sauer2024}. Furthermore, we propose two crucial modifications to the model architecture to remove and reduce artefacts caused by the \ac{erp} image format. We first give an overview of our \ac{ldm}, detailing the structure, training and inference processes. We then discuss the role of the \ac{lae} and our developments to remove the \ac{erp} border seam. Finally, we describe the function of the \ac{ddpm} and our proposed PanoDiT that reduces the \ac{erp} distortions.

\begin{figure*}[t]
\centering
\begin{subfigure}{0.33\linewidth}
    \centering
    \centerline{\includegraphics[width=\textwidth]{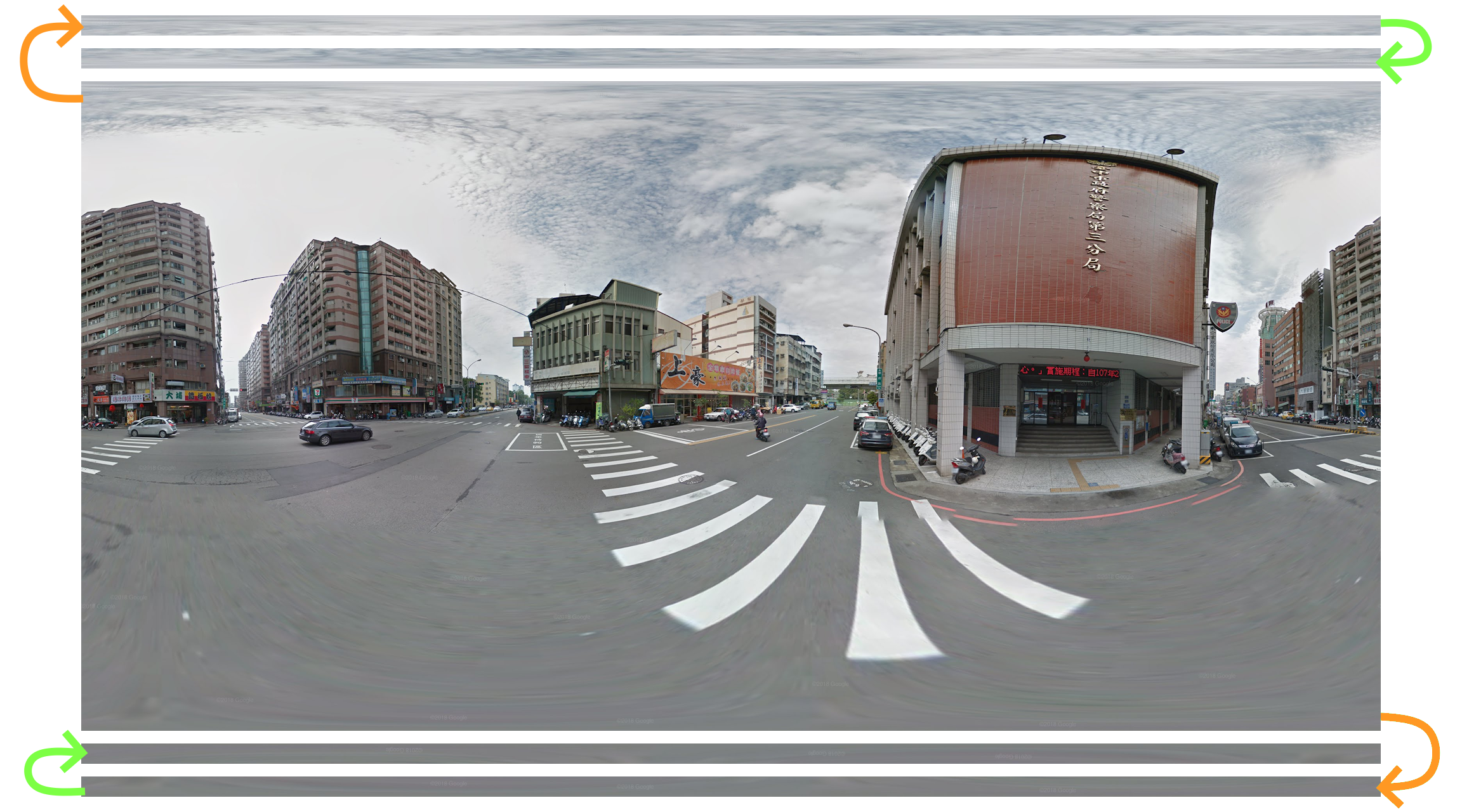}}
    \caption{Top \& Bottom Pad}\medskip
    \label{fig:ERP_pad-a}
\end{subfigure}
\hfill
\begin{subfigure}{0.33\linewidth}
    \centering
    \centerline{\includegraphics[width=\textwidth]{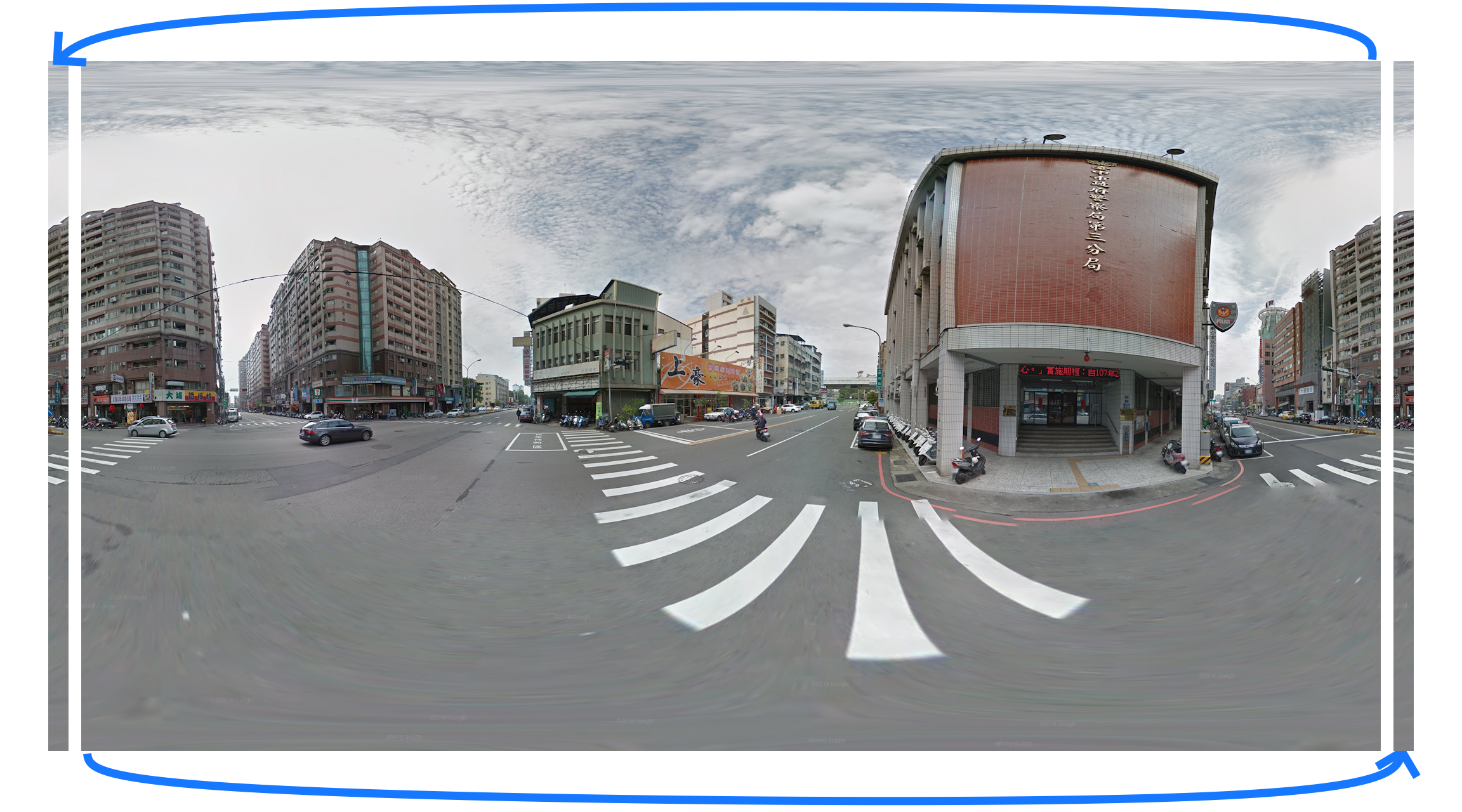}}
    \caption{Circular Pad}\medskip
    \label{fig:ERP_pad-b}
\end{subfigure}
\hfill
\begin{subfigure}{0.32\linewidth}
    \centering
    \centerline{\includegraphics[width=\textwidth]{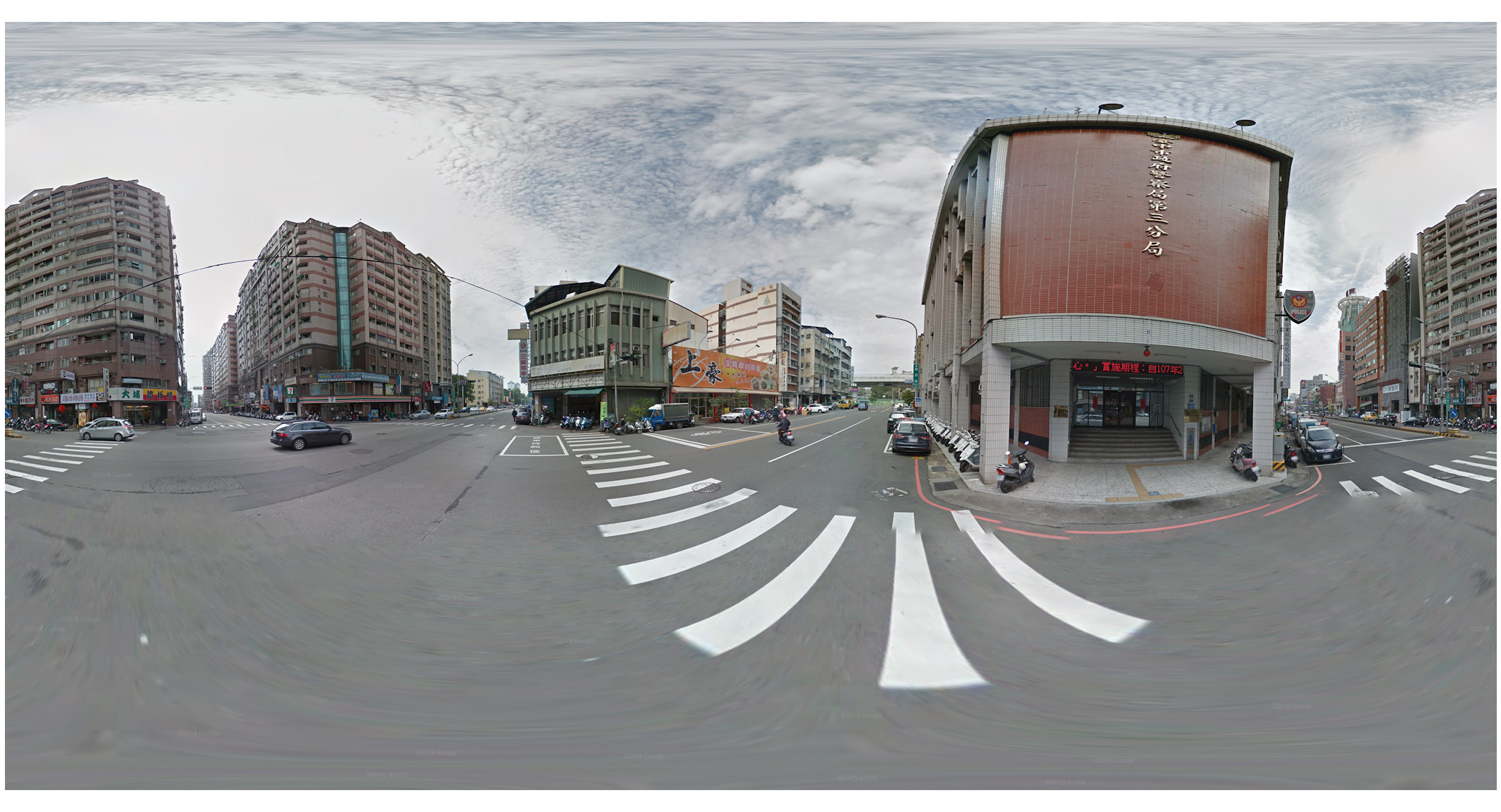}}
    \caption{Padded Image}\medskip
    \label{fig:ERP_pad-c}
\end{subfigure}
\caption{The \acs{erp} padding we use in our convolutions of the \acs{lae}. Our network uses a padding of 1 during convolutions, here we choose a padding of 8 so it can be visualised.
}
\label{fig:AE_ERP_padding}
\end{figure*}

\subsection{Model Overview}\label{LDM:methodology:subsec1}

The \ac{ldm} consists of two core components: the \ac{lae} and the \ac{ddpm}. The model's training is two-fold: first, \ac{ldr} and \ac{hdr} autoencoders are trained to encode and decode an image into and out of latent space. Second, the model learns to denoise an encoded \ac{hdri} with added noise, given an encoded \ac{lfov} \ac{ldri} and a mask. Noise is incrementally diffused onto the encoded \ac{hdri} and the network is trained to learn the previous step in the diffusion process. During inference, the model iteratively denoises random noise so that it can infer the most likely \ac{erp} \ac{hdr} environment map, given the encoded \ac{lfov} \ac{ldri} and mask as conditions. The overall pipeline for training and inference can be seen in Figure \ref{fig:model}.

\subsection{Latent Autoencoder}

The \ac{lae}, composed of an encoder $\varepsilon(.)$ and decoder $D(.)$, encodes an image $x_0$ into latent space $z_0=\varepsilon(x_0)$ and reconstructs it from this latent dimension $x_0=D(z_0)$. We use two \ac{lae}s in our model, one for \ac{ldri} $\varepsilon_{L}$ and the other for \ac{hdri} $\varepsilon_{H}$. We use the Kullback Leibler (KL) divergence autoencoder, as suggested by Esser \etal \cite{Esser2024}. Our autoencoders encode images of resolution 256x512x3 to a latent space of 32x64x4. We found that this scale is most appropriate, as the \ac{fov} masks would dramatically warp and become misaligned due to interpolation errors at lower resolution scales. We adapt our autoencoder to understand the border adjacency of \ac{erp} images by implementing an \ac{erp} padding instead of the standard convolutional padding. Our \ac{erp} padding, \Cref{fig:AE_ERP_padding}, combines circular padding at the sides of the image with a flipped mirror padding at the top and bottom to account for the adjacent pixels at the poles of \ac{erp}. This form of convolutional padding allows our autoencoders to recreate images without border seams.

\begin{figure}
    \centering
    \includegraphics[width=0.95\linewidth]{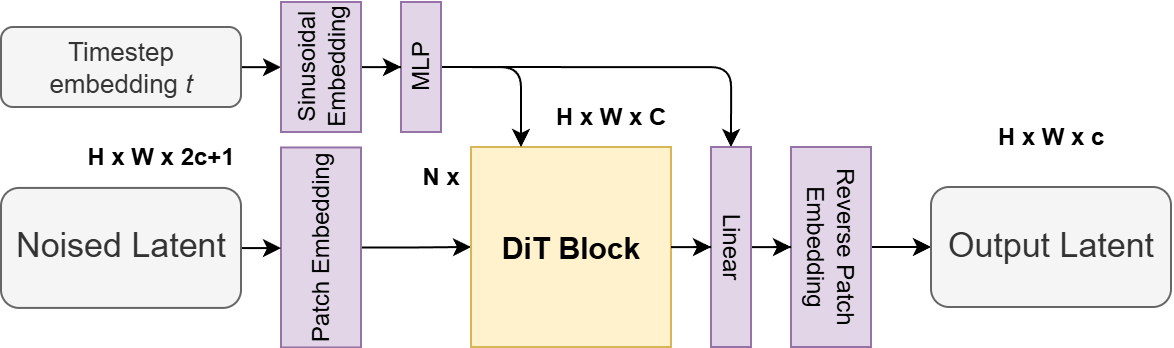}
\caption{The \ac{dit} network diagram. The input is the concatenated mask, the latent space masked image and the latent space noisy image. The output is the denoised latent space image.}
\label{fig:ditnet_diagram}
\end{figure}

\begin{figure}
    \centering
    \includegraphics[width=0.6\linewidth]{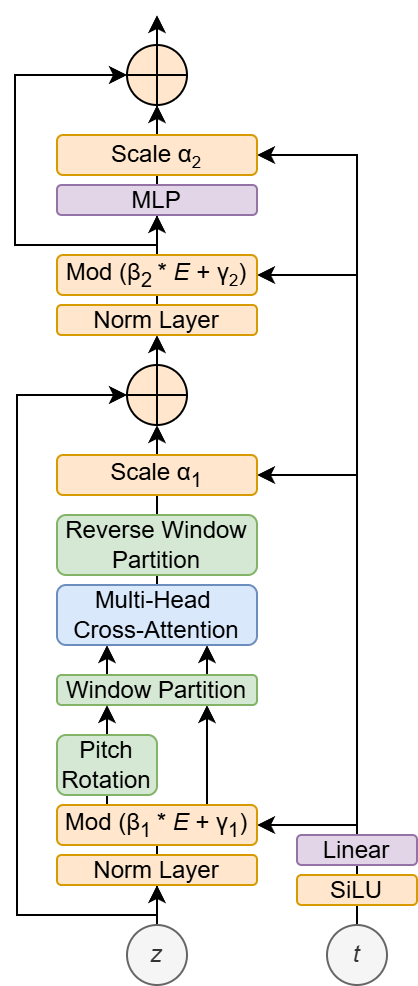}
    \caption{DPAM block diagram with added timestep embedding.}
    \label{fig:PDiT_layers}
\end{figure}

\subsection{\acl{ddpm}}

The \ac{ddpm} $\epsilon_\theta$ learns to predict the next step in the inverse distribution given the noisy latent $z_t$ and encoded conditions $z_L$ and mask $m$
\begin{equation}
\epsilon_\theta(z_{t},t,z_L,m)=(z_t-\bar{\alpha_t}z_\theta(z_t,t)),
\end{equation}
where $z_\theta$ is the estimated starting latent based on the current step $z_t$. The noisy latent $z_t$ is produced by adding noise in a Markovian manner
\begin{equation}
q(z_t|z_0)=\mathcal{N}(z_t;\sqrt{\bar{\alpha_t}}z_0,(1-\bar{\alpha_t}{\bf I}),
\end{equation}
and the constant $\bar{\alpha_t}$ is a hyperparameter. The overall model can be defined as:
\begin{equation}
L_{LDM}:=\mathbb{E}_{\varepsilon_H(x),z_L,m,\epsilon~\mathcal{N}(0,1),t}=[\parallel\epsilon-\epsilon_\theta(z_t,z_L,m,t)\parallel^2_2].
\end{equation}

\begin{table*}[t]
\centering
\begin{tabular}{@{}lccccc@{}}
\toprule
Method & si-RMSE$\downarrow$ & RMSE$\downarrow$ & RGB ang.$\downarrow$ & PSNR$\uparrow$ & FID$\downarrow$\\
\midrule
\multicolumn{6}{@{}l@{}}{INDOOR METHODS}\\
\midrule
Ours$_{U-Net}$&{0.048}&\underline{0.091}&{11.13$^\circ$}&\underline{12.98}&\underline{64.31}\\
Ours\textsubscript{PanoDiT}&\underline{0.046}&{\bf 0.085}&{10.78$^\circ$}&{\bf 13.49}&{83.49}\\
360U-Former\cite{Hilliard2024}&{\bf 0.033}&{0.110}&6.11$^\circ$&11.68&119.91\\
EverLight\cite{Dastjerdi2023}&0.087&0.239&5.75$^\circ$&10.04&{65.50}\\
Weber'22\cite{Weber2022}&{0.079}&{0.196}&\underline{4.08$^\circ$}&{12.95}&130.13\\
EMLight\cite{zhan2021emlight}&0.099&0.232&{\bf 3.99$^\circ$}&10.34&135.97\\
ImmerseGAN\cite{Dastjerdi2022}&0.091&0.215&7.89$^\circ$&10.87&{\bf 55.46}\\
\midrule
\multicolumn{6}{@{}l@{}}{OUTDOOR METHODS}\\
\midrule
Ours$_{U-Net}$&\underline{0.052}&\underline{0.152}&\underline{5.44$^\circ$}&{11.23}&{83.17}\\
Ours\textsubscript{PanoDiT}&\underline{0.052}&{\bf 0.135}&{5.62$^\circ$}&\underline{12.44}&90.74\\
360U-Former\cite{Hilliard2024}&{\bf 0.049}&{0.161}&{\bf 4.00$^\circ$}&{\bf 13.27}&102.63\\
EverLight\cite{Dastjerdi2023}&{0.162}&0.385&{8.30$^\circ$}&{11.01}&\underline{61.49}\\
ImmerseGAN\cite{Dastjerdi2022}&0.175&{0.341}&9.56$^\circ$&10.91&{\bf 34.43}\\
\bottomrule
\end{tabular}
\caption{Indoor and outdoor quantitative comparison with various illumination estimation methods. The metrics si-RMSE, RMSE, RGB ang. and PSNR are evaluated by rendering a diffuse scene. The FID score is calculated on the generated environment maps. The \textbf{best} and \underline{second-best} scores for each metric and domain are highlighted.}
\label{T_Indoor_LDM}

\end{table*}

For the U-Net \ac{ddpm} network, we use the configuration from \cite{Rombach2022}. Furthermore, we propose PanoDiT, a \ac{dit} style architecture \cite{Peebles2023} that consists of a series of \ac{vit} blocks with additional timestep embedding and can be seen in \Cref{fig:ditnet_diagram}. The purpose of PanoDiT is to reduce the distortions caused by representing the 360$^\circ$ lighting conditions of a scene at a point on a 2D \ac{erp} image. To create a network that understands the distortions that occur in the \ac{erp} format, we introduce a Diffusion Pitch Attention Module (DPAM) that combines timestep embedding with the PAM from PanoSWIN \cite{Ling2023}. We implement the timestep embedding using a shared adaLN-single parameter \cite{Perez2018}, in line with \cite{Peebles2023, Chen2024, Esser2024}. The PAM applies a window partition to the \ac{erp} image and performs cross-attention between each window with the respective window from the \ac{erp} rotated and pitched by 90$^\circ$. This allows the network to understand the relationship between objects at different positions in the \ac{erp}. The DPAM is presented in \Cref{fig:PDiT_layers}. The PanoDiT consists of 21 \ac{dit} blocks with 16 attention heads and a hidden dimension size of 1024. We use the PAM every 7th block, 3 times in total.

\section{Evaluation}\label{LDM:evaluation}

We compare our models against state-of-the-art illumination estimation methods, the results of which have been generously contributed to the community by Karimi \etal \cite{Dastjerdi2023} and can be found on the EverLight project website\footnote{\url{https://lvsn.github.io/everlight/}}. Next, we conduct an ablation study on the \ac{lae} to demonstrate the effectiveness of our \ac{erp} convolutional padding. An ablation study is not performed on the \ac{ddpm} networks as we evaluate both models when comparing against other illumination estimation methods. Additional results are included in the supplementary material.

\subsection{Implementation Details}\label{LDM:evaluation:subsec1}
All networks are trained on indoor and outdoor data at an image resolution of 256x512. For the indoor data, we use the Structured3D \cite{Zheng2019} synthetic dataset which contains 21,843 photo-realistic \ac{erp}s. For the outdoor data, we use the 360 Sun Positions \cite{Chang2018} dataset which contains 19,093 360$^\circ$ street view images of both urban roads and rural environments. We generate the \ac{hdr} pairs for both datasets using the \ac{ldr}-to-\ac{hdr} network from \cite{Hilliard2024}. We augment both datasets by horizontally rotating each panorama 4 times randomly between 60$^\circ$ and 285$^\circ$ at intervals of 45$^\circ$ with a 20\% chance of being horizontally flipped for a total of 204,680 images. When training our \ac{ldr} autoencoder, we randomly mask 50$\%$ of the panoramas with a random \ac{lfov} mask so that it can be used for both \ac{ldr} \ac{erp}s and \ac{lfov} \ac{ldri}. We split each augmented dataset into a train/test ratio of 99:1 and ensure that all augmented versions of each pair exist only in the train or the test subset to prevent over-fitting. Each \ac{ddpm} architecture is trained on a range of randomly chosen \ac{lfov} sizes $\nobreak \{40^\circ, 60^\circ, 90^\circ, 120^\circ\}$ for the masked input.
Each latent autoencoder is fine-tuned for 30 epochs on 2 NVIDIA A100 GPUs with ADAM optimiser with a learning rate of $4.5\times10^{-6}$ and betas 0.5 and 0.9. The \ac{ldr} autoencoder is fine-tuned from the f=8, KL autoencoder provided by \cite{Rombach2022}. The \ac{hdr} autoencoder is fine-tuned on our \ac{ldr} autoencoder. Both \ac{ddpm}s are trained from scratch for 50 epochs on an A100 NVIDIA GPU with the AdamW optimiser with a learning rate of $2.0\times10^{-6}$. We train our \ac{ddpm} with 1000 timesteps and use 50 for inference.

\subsection{Comparison with Illumination Estimation Methods}\label{LDM:evaluation:subsec2}

This comparison is split into indoor and outdoor scene domains, as the majority of previous methods are designed for either domain and each evaluation uses different datasets. For each domain, we compare quantitatively and qualitatively. For the quantitative measurements, \Cref{T_Indoor_LDM}, we compare semantic similarity and diffuse light position and colour accuracy \cite{Weber2022}. For our qualitative comparisons, we visually compare key locations of the generated \ac{erp}s to identify potential artefacts and assess the quality at these areas. We present environment maps horizontally rotated by 180$^\circ$ to observe potential border seams or discontinuous generation. The top and bottom faces of the cube map representation are used to identify distortions at the poles. 


\subsubsection{Indoor Scenes}\label{LDM:evaluation:subsec2:subsubsec1}

For indoor scenes, we follow the two-fold evaluation protocol of \cite{Weber2022}: first, the generated environment maps are used to light a scene with 9 diffuse spheres on a ground plane. This measures the ability of each model to produce accurate light positions, colours and intensities on a diffuse surface. Second, the plausibility of the generated \ac{hdri} is measured with \ac{fid} score. The rendering tests use the 224 images from the test split of the Laval HDR Indoor dataset \cite{Gardner2017}. The \ac{fid} score uses the 305 images from the Laval HDR Indoor test set \cite{Gardner2017} and the 192 indoor images from \cite{Cheng2018}. Additional datasets help to remove the potential bias of using a single dataset. For each metric, we extract 10 views of 50$^\circ$ for each image. 

The \ac{fid} score of our U-Net model outperforms all models by a noticeable margin, except for EverLight and ImmerseGAN. This is supported by our environment maps, presented in \Cref{fig:LDM_indoor_renders}, which show more complex and detailed scenes with greater variance than 360U-Former and Weber \etal. In terms of \ac{erp} artefacts, our U-Net and PanoDiT \ac{ldm}s and the 360U-Former do not introduce the border seam that is visible in \cite{Weber2022} and subtly perceptible in EverLight. When comparing generations at the poles, all models produce an unrealistic shadowing effect, a result of a light source generated on one side of the panorama and not on the other. This pinching effect is subtly present in our PanoDiT model, proving that the DPAM can reduce \ac{erp} artefacts at the poles. In terms of lighting accuracy, our PanoDiT model performs at state-of-the-art level at estimating the colours of the light sources, measured by si-RMSE, RMSE and PSNR. However, the RGB angular error is particularly poor for indoor scenes.

\subsubsection{Outdoor Scenes}\label{LDM:evaluation:subsec2:subsubsec2}

For outdoor scenes, the quantitative comparison is conducted similarly to the indoor scenes, the input is changed to 3 perspective crops at azimuth spacing \{0, 120, 240\} of 90$^\circ$ \ac{fov}. We use the 893 outdoor panoramas from the \cite{Cheng2018} dataset, totalling 2,517 images for evaluation. We compare our approach with the works of 
Everlight \cite{Dastjerdi2023}, ImmerseGAN \cite{Dastjerdi2022} and 360U-Former \cite{Hilliard2024}.

\begin{figure*}
\footnotesize
\centering
\setlength\tabcolsep{0.6pt}
\renewcommand{\arraystretch}{0.1}
\resizebox{\linewidth}{!}{\begin{tabularx}{\linewidth}{m{0.025\linewidth}ccccc}

\rotatebox{90}{\textbf{Input}}  &\raisebox{-0.1\height}{\centering \includegraphics[width=0.1\textwidth,valign=m]{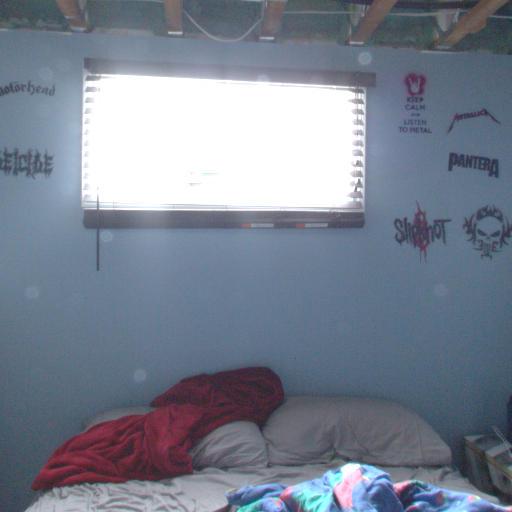}} &\raisebox{-0.1\height}{\centering \includegraphics[width=0.1\textwidth,valign=m]{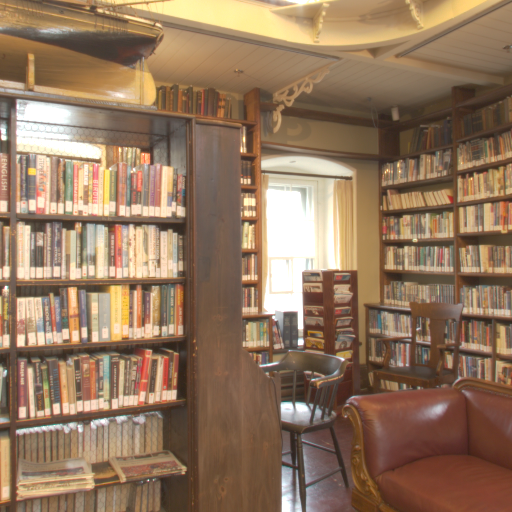}} &\raisebox{-0.1\height}{\centering \includegraphics[width=0.1\textwidth,valign=m]{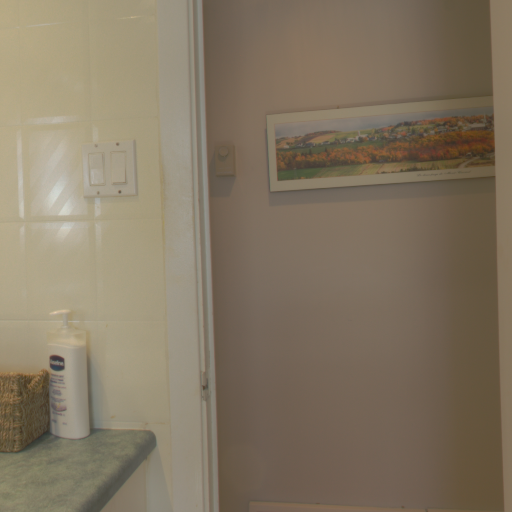}} &\raisebox{-0.1\height}{\centering \includegraphics[width=0.1\textwidth,valign=m]{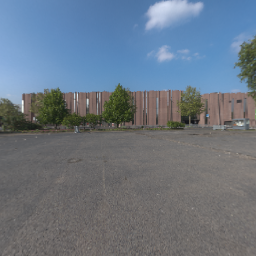}} &\raisebox{-0.1\height}{\centering \includegraphics[width=0.1\textwidth,valign=m]{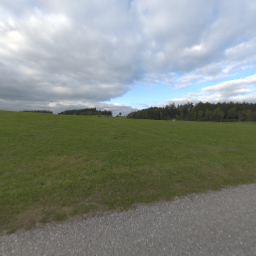}}\\

& \rule{0pt}{0.1pt} & \rule{0pt}{0.1pt} & \rule{0pt}{0.1pt} & \rule{0pt}{0.1pt} & \rule{0pt}{0.1pt} \\

\multirow{3}{*}{\rotatebox{90}{\textbf{Ground Truth}}} &\raisebox{-0.2\height}{\includegraphics[width=0.19\textwidth,valign=m]{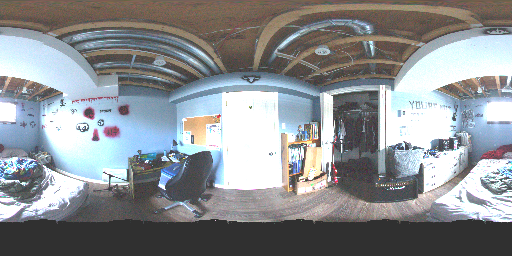}} &\raisebox{-0.2\height}{\includegraphics[width=0.19\textwidth,valign=m]{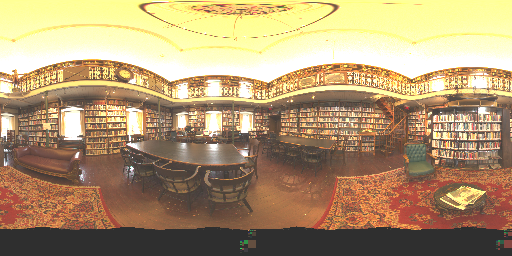}} &\raisebox{-0.2\height}{\includegraphics[width=0.19\textwidth,valign=m]{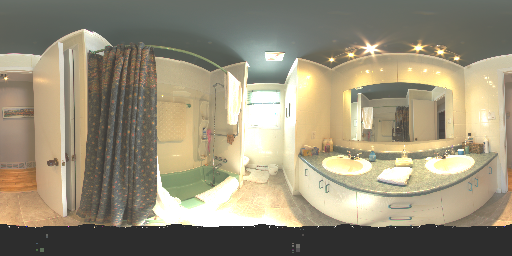}}&\raisebox{-0.2\height}{\includegraphics[width=0.19\textwidth,valign=m]{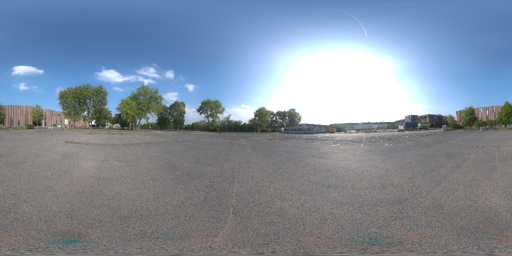}} &\raisebox{-0.2\height}{\includegraphics[width=0.19\textwidth,valign=m]{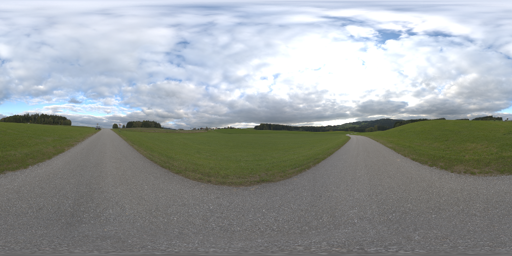}}\\

& \rule{0pt}{0.1pt} & \rule{0pt}{0.1pt} & \rule{0pt}{0.1pt} & \rule{0pt}{0.1pt} & \rule{0pt}{0.1pt} \\

&\includegraphics[width=0.19\textwidth,valign=m]{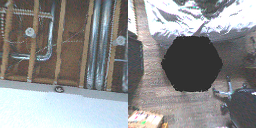} &\includegraphics[width=0.19\textwidth,valign=m]{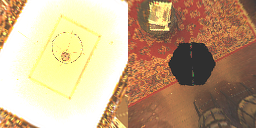} &\includegraphics[width=0.19\textwidth,valign=m]{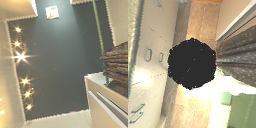}&\includegraphics[width=0.19\textwidth,valign=m]{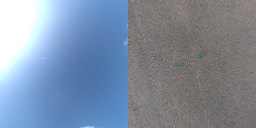} &\includegraphics[width=0.19\textwidth,valign=m]{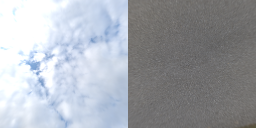}\\

& \rule{0pt}{0.1pt} & \rule{0pt}{0.1pt} & \rule{0pt}{0.1pt} & \rule{0pt}{0.1pt} & \rule{0pt}{0.1pt} \\

\multirow{3}{*}{\rotatebox{90}{\textbf{Weber '22 \cite{Weber2022}}}} &\raisebox{-0.3\height}{\includegraphics[width=0.19\textwidth,valign=m]{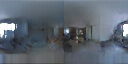}} &\raisebox{-0.3\height}{\includegraphics[width=0.19\textwidth,valign=m]{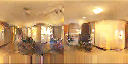}} &\raisebox{-0.3\height}{\includegraphics[width=0.19\textwidth,valign=m]{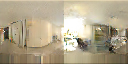}}\\

& \rule{0pt}{0.1pt} & \rule{0pt}{0.1pt} & \rule{0pt}{0.1pt} & \rule{0pt}{0.1pt} & \rule{0pt}{0.1pt} \\

&\includegraphics[width=0.19\textwidth,valign=m]{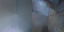} &\includegraphics[width=0.19\textwidth,valign=m]{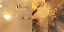} &\includegraphics[width=0.19\textwidth,valign=m]{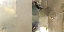}\\

& \rule{0pt}{0.1pt} & \rule{0pt}{0.1pt} & \rule{0pt}{0.1pt} & \rule{0pt}{0.1pt} & \rule{0pt}{0.1pt} \\

\multirow{3}{*}{\rotatebox{90}{\textbf{EverLight \cite{Dastjerdi2023}}}} &\raisebox{-0.25\height}{\includegraphics[width=0.19\textwidth,valign=m]{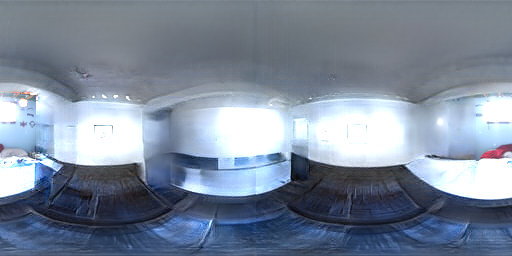}} &\raisebox{-0.25\height}{\includegraphics[width=0.19\textwidth,valign=m]{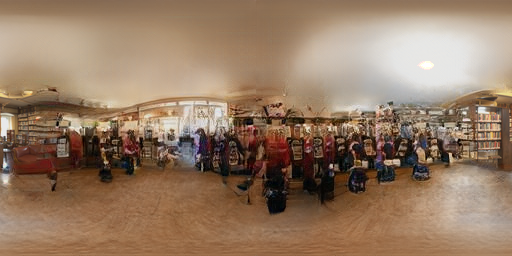}} &\raisebox{-0.25\height}{\includegraphics[width=0.19\textwidth,valign=m]{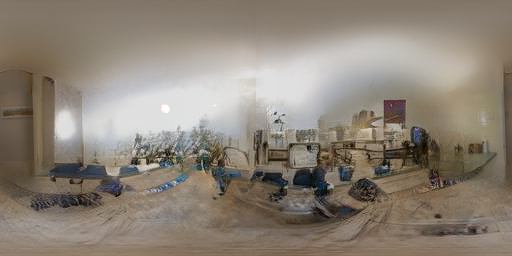}} &\raisebox{-0.25\height}{\includegraphics[width=0.19\textwidth,valign=m]{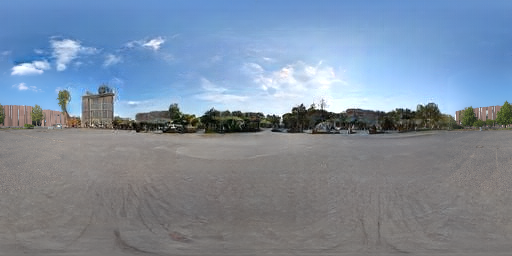}} &\raisebox{-0.25\height}{\includegraphics[width=0.19\textwidth,valign=m]{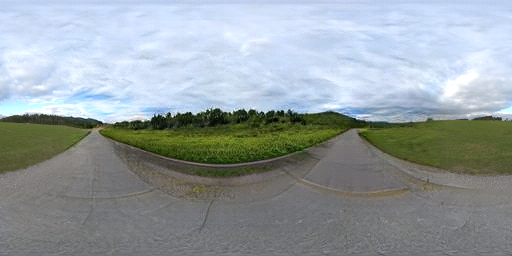}}\\

& \rule{0pt}{0.1pt} & \rule{0pt}{0.1pt} & \rule{0pt}{0.1pt} & \rule{0pt}{0.1pt} & \rule{0pt}{0.1pt} \\

&\includegraphics[width=0.19\textwidth,valign=m]{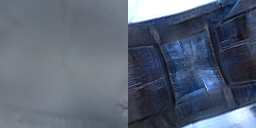} &\includegraphics[width=0.19\textwidth,valign=m]{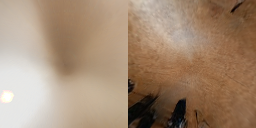} &\includegraphics[width=0.19\textwidth,valign=m]{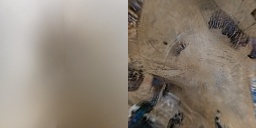} &\includegraphics[width=0.19\textwidth,valign=m]{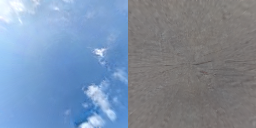} &\includegraphics[width=0.19\textwidth,valign=m]{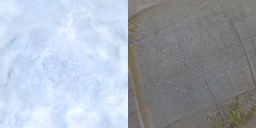}\\

& \rule{0pt}{0.1pt} & \rule{0pt}{0.1pt} & \rule{0pt}{0.1pt} & \rule{0pt}{0.1pt} & \rule{0pt}{0.1pt} \\

\multirow{3}{*}{\rotatebox{90}{\textbf{360U-Former\cite{Hilliard2024}}}}
&\raisebox{-0.4\height}{\includegraphics[width=0.19\textwidth,valign=m]{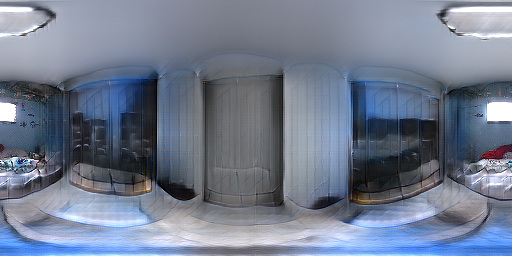}} &\raisebox{-0.4\height}{\includegraphics[width=0.19\textwidth,valign=m]{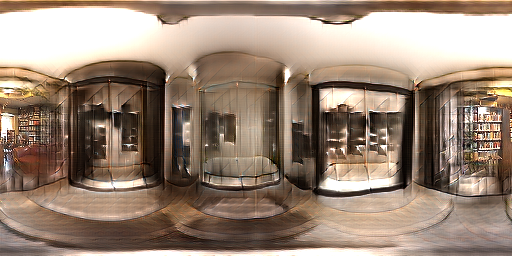}} &\raisebox{-0.4\height}{\includegraphics[width=0.19\textwidth,valign=m]{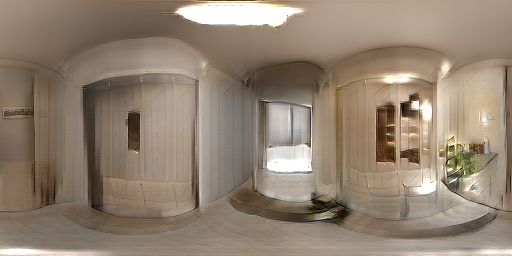}} &\raisebox{-0.4\height}{\includegraphics[width=0.19\textwidth,valign=m]{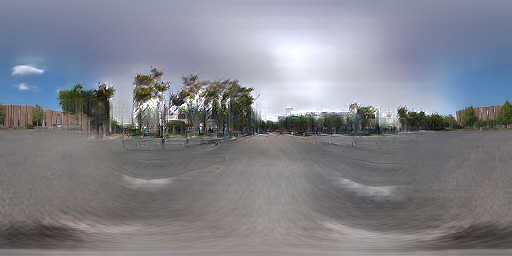}} &\raisebox{-0.4\height}{\includegraphics[width=0.19\textwidth,valign=m]{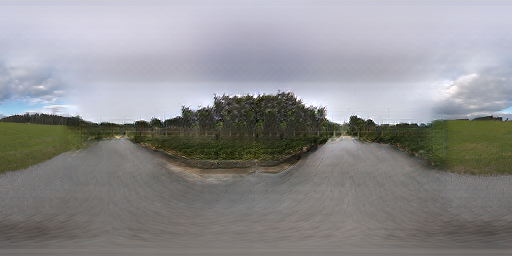}}\\

& \rule{0pt}{0.1pt} & \rule{0pt}{0.1pt} & \rule{0pt}{0.1pt} & \rule{0pt}{0.1pt} & \rule{0pt}{0.1pt} \\

&\includegraphics[width=0.19\textwidth,valign=m]{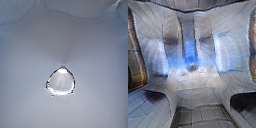} &\includegraphics[width=0.19\textwidth,valign=m]{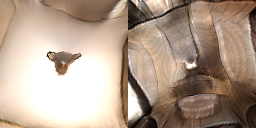} &\includegraphics[width=0.19\textwidth,valign=m]{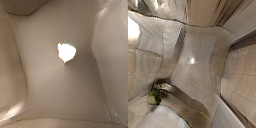} &\includegraphics[width=0.19\textwidth,valign=m]{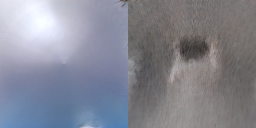} &\includegraphics[width=0.19\textwidth,valign=m]{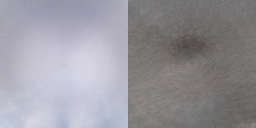}\\

& \rule{0pt}{0.1pt} & \rule{0pt}{0.1pt} & \rule{0pt}{0.1pt} & \rule{0pt}{0.1pt} & \rule{0pt}{0.1pt} \\

\multirow{3}{*}{\rotatebox{90}{\textbf{Ours (U-Net)}}} &\raisebox{-0.2\height}{\includegraphics[width=0.19\textwidth,valign=m]{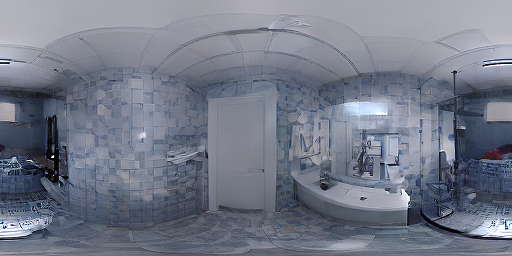}} &\raisebox{-0.2\height}{\includegraphics[width=0.19\textwidth,valign=m]{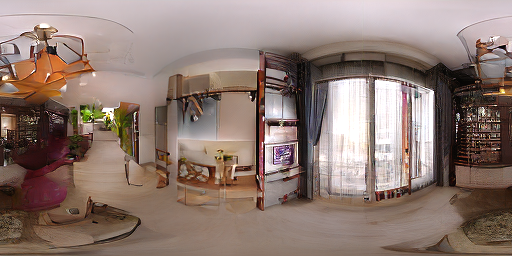}} &\raisebox{-0.2\height}{\includegraphics[width=0.19\textwidth,valign=m]{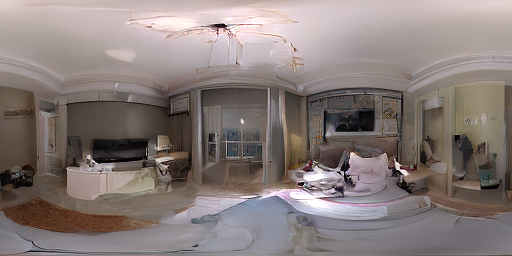}} &\raisebox{-0.2\height}{\includegraphics[width=0.19\textwidth,valign=m]{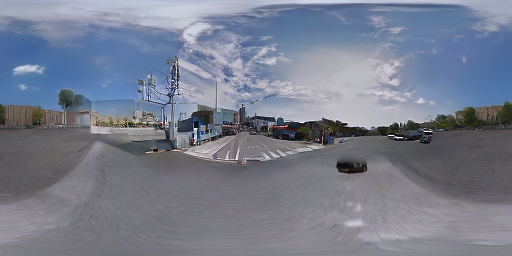}} &\raisebox{-0.2\height}{\includegraphics[width=0.19\textwidth,valign=m]{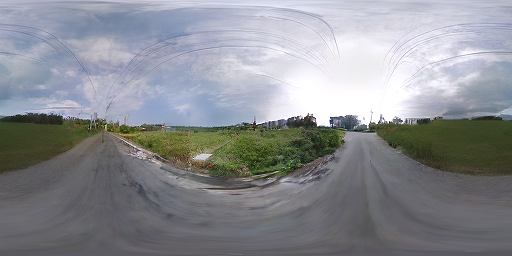}}\\

& \rule{0pt}{0.1pt} & \rule{0pt}{0.1pt} & \rule{0pt}{0.1pt} & \rule{0pt}{0.1pt} & \rule{0pt}{0.1pt} \\

&\includegraphics[width=0.19\textwidth,valign=m]{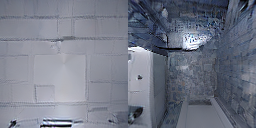} &\includegraphics[width=0.19\textwidth,valign=m]{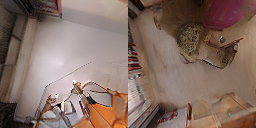} &\includegraphics[width=0.19\textwidth,valign=m]{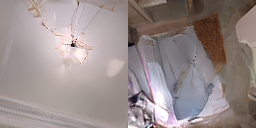} &\includegraphics[width=0.19\textwidth,valign=m]{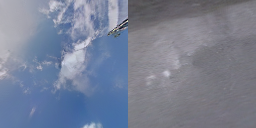} &\includegraphics[width=0.19\textwidth,valign=m]{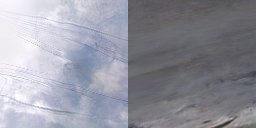}\\

& \rule{0pt}{0.1pt} & \rule{0pt}{0.1pt} & \rule{0pt}{0.1pt} & \rule{0pt}{0.1pt} & \rule{0pt}{0.1pt} \\

\multirow{3}{*}{\rotatebox{90}{\textbf{Ours (PanoDiT)}}} &\raisebox{-0.4\height}{\includegraphics[width=0.19\textwidth,valign=m]{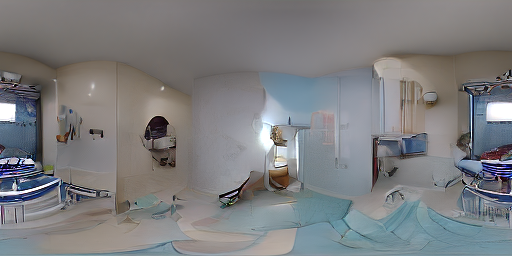}} &\raisebox{-0.4\height}{\includegraphics[width=0.19\textwidth,valign=m]{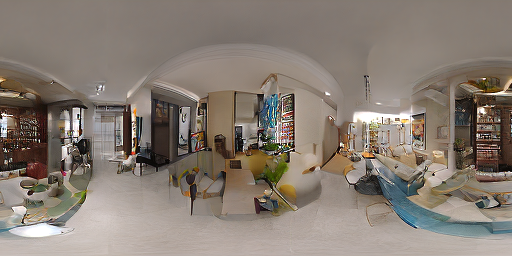}} &\raisebox{-0.4\height}{\includegraphics[width=0.19\textwidth,valign=m]{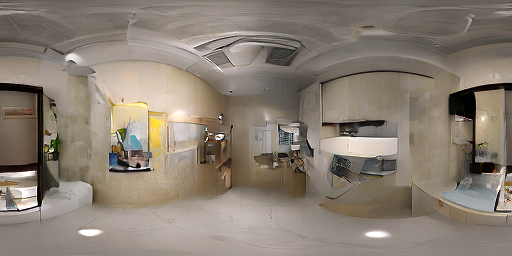}} &\raisebox{-0.4\height}{\includegraphics[width=0.19\textwidth,valign=m]{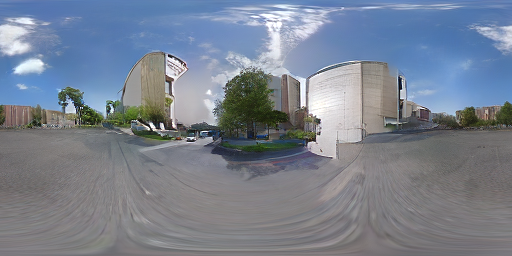}} &\raisebox{-0.4\height}{\includegraphics[width=0.19\textwidth,valign=m]{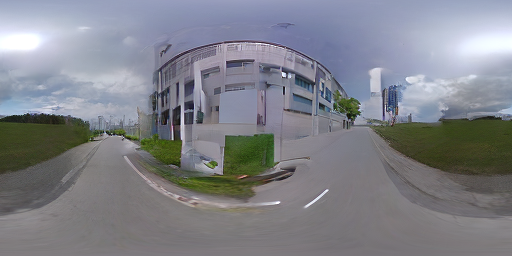}}\\

& \rule{0pt}{0.1pt} & \rule{0pt}{0.1pt} & \rule{0pt}{0.1pt} & \rule{0pt}{0.1pt} & \rule{0pt}{0.1pt} \\

&\includegraphics[width=0.19\textwidth,valign=m]{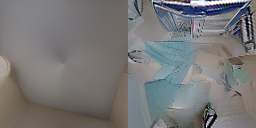} &\includegraphics[width=0.19\textwidth,valign=m]{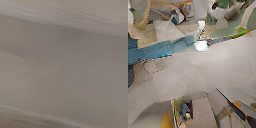} &\includegraphics[width=0.19\textwidth,valign=m]{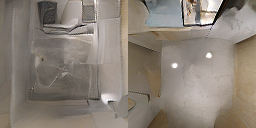} &\includegraphics[width=0.19\textwidth,valign=m]{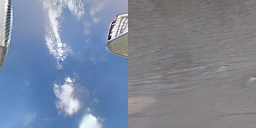} &\includegraphics[width=0.19\textwidth,valign=m]{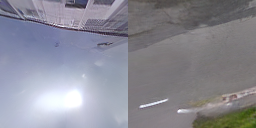}
\end{tabularx}}
\vspace{-1em}
\caption{\small Indoor qualitative comparison of our method with state-of-the-art methods. For each method and input \acs{lfov} image, we show the \acs{erp} rotated 180$^\circ$, to show potential border seams, and the top and bottom cube map faces, to visualise generation at the poles.}
\label{fig:LDM_indoor_renders}
\end{figure*}

The results in \Cref{T_Indoor_LDM} show that our approaches outperform EverLight and ImmerseGAN and are competitive with 360U-Former in terms of RMSE and si-RMSE. When observing the environment maps in \Cref{fig:LDM_indoor_renders}, our models consistently and correctly estimate the sun position and the amount of cloud cover. Supported by the \ac{fid} score, our method can generate realistic features and textures compared to the 360U-Former. However, EverLight and ImmerseGAN continue to outperform our approach, potentially due to training on a larger dataset. 
In terms of lighting accuracy, the PanoDiT model outperforms the U-Net. Additionally, there is a stronger correlation towards PanoDiT on the indoor datasets, which have more complex lighting with multiple sources to estimate. This could suggest that an inherent understanding of the \ac{erp} warping could assist in accurate light positioning.



\begin{figure*}[t]
\begin{center}
\setlength\tabcolsep{1pt}
\renewcommand{\arraystretch}{0.05}
\def\colW{0.17}
\begin{tabularx}{0.88\linewidth}{lXXXXX}
    {\rotatebox{90}{\textbf{GT}}} &\raisebox{-0.4\height}{\includegraphics[width=\colW\textwidth]{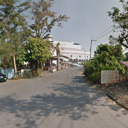}} 
    &\raisebox{-0.4\height}{\includegraphics[width=\colW\textwidth]{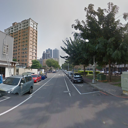}} &\raisebox{-0.4\height}{\includegraphics[width=\colW\textwidth]{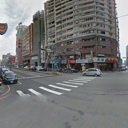}} &\raisebox{-0.4\height}{\includegraphics[width=\colW\textwidth]{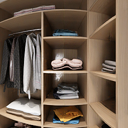}} &\raisebox{-0.4\height}{\includegraphics[width=\colW\textwidth]{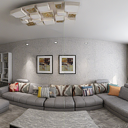}}\\

    \rule{0pt}{0.1pt} & \rule{0pt}{0.1pt} & \rule{0pt}{0.1pt} & \rule{0pt}{0.1pt}\\
    
    {\rotatebox{90}{\textbf{Ours}}} &\raisebox{-0.3\height}{\includegraphics[width=\colW\textwidth]{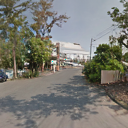}} 
    &\raisebox{-0.3\height}{\includegraphics[width=\colW\textwidth]{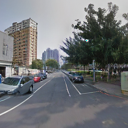}} &\raisebox{-0.3\height}{\includegraphics[width=\colW\textwidth]{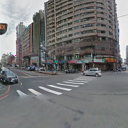} }&\raisebox{-0.3\height}{\includegraphics[width=\colW\textwidth]{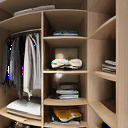} }&\raisebox{-0.3\height}{\includegraphics[width=\colW\textwidth]{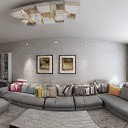}}\\

    \rule{0pt}{0.1pt} & \rule{0pt}{0.1pt} & \rule{0pt}{0.1pt} & \rule{0pt}{0.1pt} & \rule{0pt}{0.1pt}\\
    
    {\rotatebox{90}{\textbf{Standard}}} &\raisebox{-0.2\height}{\includegraphics[width=\colW\textwidth]{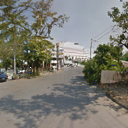}}
    &\raisebox{-0.2\height}{\includegraphics[width=\colW\textwidth]{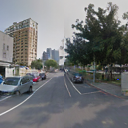} }&\raisebox{-0.2\height}{\includegraphics[width=\colW\textwidth]{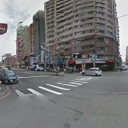} }&\raisebox{-0.2\height}{\includegraphics[width=\colW\textwidth]{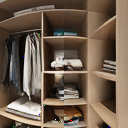}} &\raisebox{-0.2\height}{\includegraphics[width=\colW\textwidth]{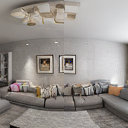}}\\
\end{tabularx}
\end{center}
\caption{Comparison of the \acs{ldr} \acs{erp}s recreated by the \ac{lae} with and without our \acs{erp} padding. A crop is taken from the centre of the panoramas to view the border seam.
}
\label{fig:ae_rotations}
\end{figure*}

\begin{table*}[t]
    \centering
    \begin{tabular}{@{}lccccc@{}}
        \toprule
        & si-RMSE$\downarrow$ & RMSE$\downarrow$ & SSIM$\uparrow$ & PSNR$\uparrow$ & FID$\downarrow$\\
        \midrule
        \ac{erp} Convolutional Padding&{\bf 0.04155}&\textbf{0.04155}&\textbf{0.8953}&\textbf{28.04}&7.62\\
        Standard Convolutional Padding&{0.04331}&{0.04335}&{0.8896}&27.60&\textbf{7.61}\\
        \bottomrule
    \end{tabular}
    \caption{A quantitative comparison of our \ac{lae}'s ability to reconstruct an \acs{ldr} \acs{erp} against an \ac{lae} with standard convolutional padding. The best results are highlighted in \textbf{bold}.}
    \label{T_Ablation_AE}
\end{table*}

\subsubsection{\ac{ddpm} Comparison}\label{LDM:evaluation:subsec2:subsubsec3}

The quantitative results in \Cref{T_Indoor_LDM} show that the U-Net, despite being a smaller and less complex model, can generate more plausible and higher-quality images than the PanoDiT. Whereas the PanoDiT more accurately estimates the lighting conditions. Comparing the panoramas in \Cref{fig:LDM_indoor_renders}, the U-Net produces panoramas that have fewer artefacts, while generated objects and features appear familiar. The PanoDiT tends to produce abstract features and textures. However, the warping at the poles of the \ac{erp} is reduced compared to the U-Net.

For both models, we occasionally observe that the generated scene does not semantically or contextually match that of the ground truth. For indoor scenes, we often notice living spaces and bathrooms being generated. For outdoor scenes, a rural space may be generated as an urban space and vice versa. A severe case of this occurs when a scene will be generated as indoor when the ground truth is outdoor and vice versa. These problems are likely a consequence of training on limited datasets, with only a small variance in scene type.

\subsection{Latent Autoencoder Ablation}\label{LDM:evaluation:subsec3}

In \Cref{fig:ae_rotations}, we demonstrate the ability of our proposed \ac{erp} convolutional padding against standard convolutional padding to remove border seam artefact by comparing \ac{erp}s reconstructed by the \ac{ldr} \ac{lae}. By simply using an \ac{lae} with \ac{erp} convolutional padding, we can train an \ac{ldm} to produce \ac{erp} images with the same features on either of the side borders without a border seam. This is further supported by a metric comparison in \Cref{T_Ablation_AE}, where our approach with \ac{erp} convolutional padding performs better on all measures except for the \ac{fid} score, which is only marginally worse.

\section{Conclusion and Future Work}
\label{LDM:Conclusion}

We have presented a novel approach to \ac{hdr} environment map estimation of a scene from a single \ac{lfov} \ac{ldri}. Our approach removes the border seam artefact by implementing an \ac{erp} convolutional padding in the \ac{lae}, simultaneously improving overall metric performance. Furthermore, we have proposed a novel \ac{dit} architecture designed to better understand the \ac{erp} image format. The PanoDiT reduces the artefacts produced at the poles of the panorama and can more accurately estimate the lighting conditions than the U-Net model. This comes at an expense to the generated image quality measured by the FID score. Evaluating with standard benchmarks, we demonstrate that both of our models compete with state-of-the-art approaches whilst training on a smaller dataset than the models that perform best in terms of FID score. Future work should investigate novel \ac{erp} architecture adaptations to ultimately remove \ac{erp} distortions.

\clearpage
\setcounter{page}{1}
\maketitlesupplementary

\begin{table*}[th]
\centering
\begin{tabular}{@{}lccccc@{}}
\toprule
Method & si-RMSE$\downarrow$ & RMSE$\downarrow$ & RGB ang.$\downarrow$ & PSNR$\uparrow$ & FID$\downarrow$\\
\midrule
\multicolumn{6}{@{}l@{}}{INDOOR METHODS}\\
\midrule
Ours$_{U-Net}$&{0.048}&\underline{0.091}&{11.13$^\circ$}&\underline{12.98}&\underline{64.31}\\
Ours\textsubscript{PanoDiT}&\underline{0.046}&{\bf 0.085}&{10.78$^\circ$}&{\bf 13.49}&{83.49}\\
360U-Former\cite{Hilliard2024}&{\bf 0.033}&{0.110}&6.11$^\circ$&11.68&119.91\\
Hilliard'23\cite{Hilliard2023}&0.112&0.300&6.50$^\circ$&10.05&158.60\\
EverLight\cite{Dastjerdi2023}&0.087&0.239&5.75$^\circ$&10.04&{65.50}\\
Weber'22\cite{Weber2022}&{0.079}&{0.196}&\underline{4.08$^\circ$}&{12.95}&130.13\\
StyleLight\cite{Wang2022stylelight}&0.130&0.261&7.05$^\circ$&{12.85}&121.60\\
Gardner'19(1)\cite{Gardner2019}&0.099&0.229&4.42$^\circ$&12.21&410.12\\
Gardner'19(3)\cite{Gardner2019}&0.105&0.507&4.59$^\circ$&10.90&386.43\\
Gardner'17\cite{Gardner2017}&0.123&0.628&8.29$^\circ$&10.22&253.40\\
Garon'19\cite{Garon2019}&0.096&0.255&8.06$^\circ$&9.73&324.51\\
Lighthouse\cite{Srinivasan2020}&0.121&0.254&4.56$^\circ$&9.81&174.52\\
EMLight\cite{zhan2021emlight}&0.099&0.232&{\bf 3.99$^\circ$}&10.34&135.97\\
EnvmapNet\cite{Somanath2021}&0.097&0.286&7.67$^\circ$&11.74&221.85 \\
ImmerseGAN\cite{Dastjerdi2022}&0.091&0.215&7.89$^\circ$&10.87&{\bf 55.46}\\
\midrule
\multicolumn{6}{@{}l@{}}{OUTDOOR METHODS}\\
\midrule
Ours$_{U-Net}$&\underline{0.052}&\underline{0.152}&\underline{5.44$^\circ$}&{11.23}&{83.17}\\
Ours\textsubscript{PanoDiT}&\underline{0.052}&{\bf 0.135}&{5.62$^\circ$}&\underline{12.44}&90.74\\
360U-Former\cite{Hilliard2024}&{\bf 0.049}&{0.161}&{\bf 4.00$^\circ$}&{\bf 13.27}&102.63\\
EverLight\cite{Dastjerdi2023}&{0.162}&0.385&{8.30$^\circ$}&{11.01}&\underline{61.49}\\

ImmerseGAN\cite{Dastjerdi2022}&0.175&{0.341}&9.56$^\circ$&10.91&{\bf 34.43}\\
Zhang '19\cite{Zhang2019LightEst} &0.225&1.058&11.80$^\circ$&10.91&449.49\\
\bottomrule
\end{tabular}
\caption{Indoor and outdoor quantitative comparison with various illumination estimation methods. The metrics si-RMSE, RMSE, RGB ang. and PSNR are evaluated by rendering a diffuse scene. The FID score is calculated on the generated environment maps. The \textbf{best} and \underline{second-best} scores for each metric and domain are highlighted.}
\label{T_extra_LDM}
\end{table*}

\section{Additional Quantitative Results}\label{LDM:1}

\Cref{T_extra_LDM} contains an extensive comparison with illumination estimation methods. Two versions of \cite{Gardner2019} are compared: the original (3) where 3 light sources are estimated, and a version (1) trained to predict a single parametric light. We also compare to Lighthouse \cite{Srinivasan2020}, which expects a stereo pair as input, instead a second image is generated with a small baseline using \cite{Wiles2020} (visual inspection confirmed this yields results comparable to the published work). For \cite{Garon2019}, the coordinates of the image centre are selected for the object position. For \cite{Somanath2021}, the proposed “Cluster ID loss”, and tonemapping are used with a pix2pixHD \cite{Wang2018Pix2pixHD} network architecture. 


\section{Additional Qualitative Results}\label{LDM:2}

\begin{figure*}[t]
\footnotesize
\centering
\setlength\tabcolsep{0.6pt}
\renewcommand{\arraystretch}{0.1}
\resizebox{\linewidth}{!}{\begin{tabularx}{\linewidth}{m{0.025\linewidth}ccccc}

\rotatebox{90}{\textbf{Input}}  &\raisebox{-0.1\height}{\centering \includegraphics[width=0.1\textwidth,valign=m]{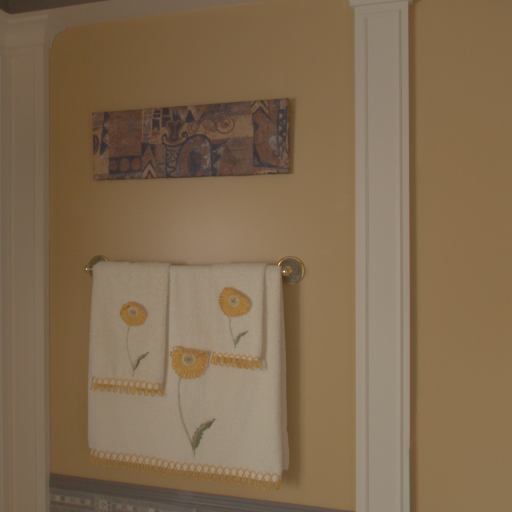}} &\raisebox{-0.1\height}{\centering \includegraphics[width=0.1\textwidth,valign=m]{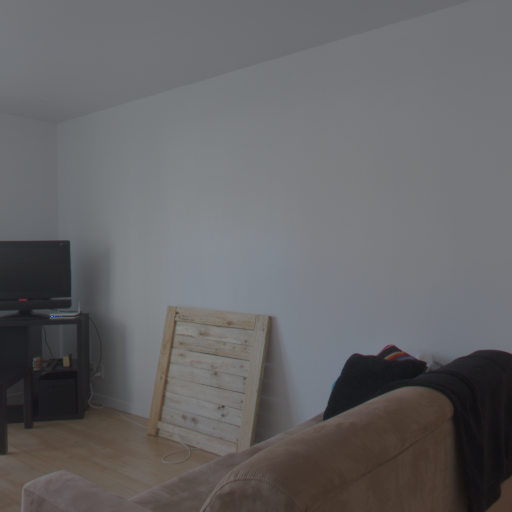}} &\raisebox{-0.1\height}{\centering \includegraphics[width=0.1\textwidth,valign=m]{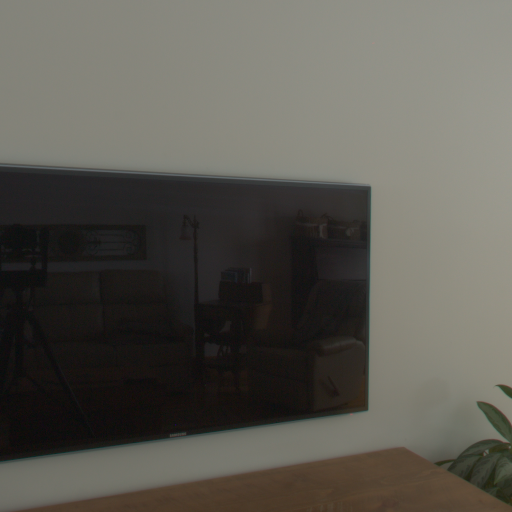}} &\raisebox{-0.1\height}{\centering \includegraphics[width=0.1\textwidth,valign=m]{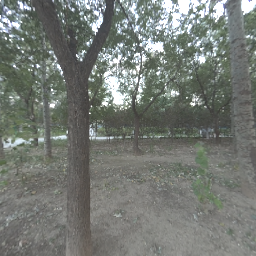}} &\raisebox{-0.1\height}{\centering \includegraphics[width=0.1\textwidth,valign=m]{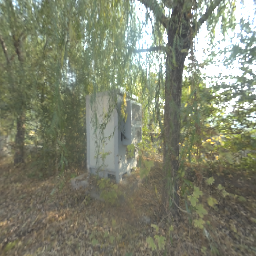}}\\

& \rule{0pt}{0.1pt} & \rule{0pt}{0.1pt} & \rule{0pt}{0.1pt} & \rule{0pt}{0.1pt} & \rule{0pt}{0.1pt} \\

\multirow{3}{*}{\rotatebox{90}{\textbf{Ground Truth}}} &\raisebox{-0.35\height}{\includegraphics[width=0.19\textwidth,valign=m]{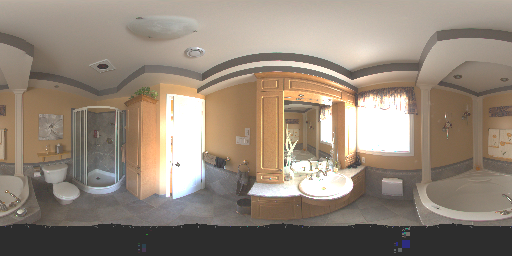}} &\raisebox{-0.35\height}{\includegraphics[width=0.19\textwidth,valign=m]{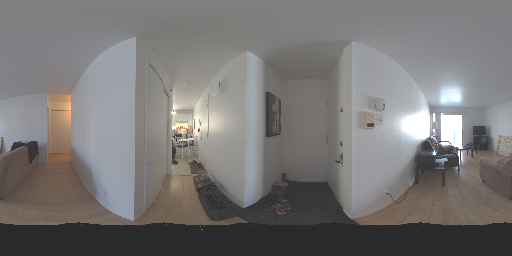}} &\raisebox{-0.35\height}{\includegraphics[width=0.19\textwidth,valign=m]{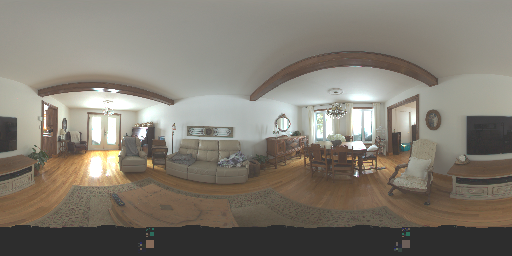}}&\raisebox{-0.35\height}{\includegraphics[width=0.19\textwidth,valign=m]{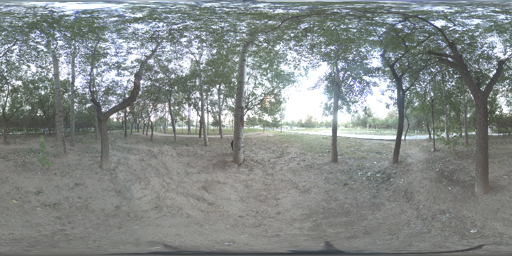}} &\raisebox{-0.35\height}{\includegraphics[width=0.19\textwidth,valign=m]{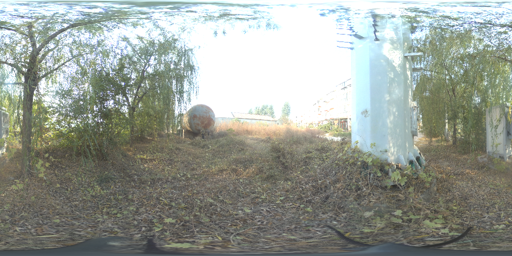}}\\

& \rule{0pt}{0.1pt} & \rule{0pt}{0.1pt} & \rule{0pt}{0.1pt} & \rule{0pt}{0.1pt} & \rule{0pt}{0.1pt} \\

&\includegraphics[width=0.19\textwidth,valign=m]{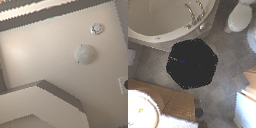} &\includegraphics[width=0.19\textwidth,valign=m]{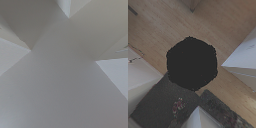} &\includegraphics[width=0.19\textwidth,valign=m]{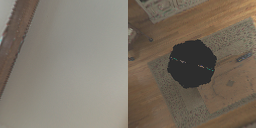}&\includegraphics[width=0.19\textwidth,valign=m]{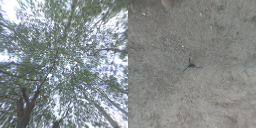} &\includegraphics[width=0.19\textwidth,valign=m]{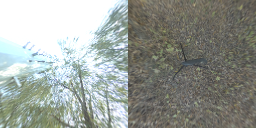}\\

& \rule{0pt}{0.1pt} & \rule{0pt}{0.1pt} & \rule{0pt}{0.1pt} & \rule{0pt}{0.1pt} & \rule{0pt}{0.1pt} \\

\multirow{3}{*}{\rotatebox{90}{\textbf{Weber '22 \cite{Weber2022}}}} &\raisebox{-0.5\height}{\includegraphics[width=0.19\textwidth,valign=m]{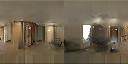}} &\raisebox{-0.5\height}{\includegraphics[width=0.19\textwidth,valign=m]{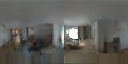}} &\raisebox{-0.5\height}{\includegraphics[width=0.19\textwidth,valign=m]{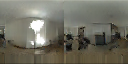}}\\

& \rule{0pt}{0.1pt} & \rule{0pt}{0.1pt} & \rule{0pt}{0.1pt} & \rule{0pt}{0.1pt} & \rule{0pt}{0.1pt} \\

&\includegraphics[width=0.19\textwidth,valign=m]{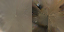} &\includegraphics[width=0.19\textwidth,valign=m]{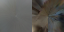} &\includegraphics[width=0.19\textwidth,valign=m]{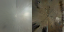}\\

& \rule{0pt}{0.1pt} & \rule{0pt}{0.1pt} & \rule{0pt}{0.1pt} & \rule{0pt}{0.1pt} & \rule{0pt}{0.1pt} \\

\multirow{3}{*}{\rotatebox{90}{\textbf{EverLight \cite{Dastjerdi2023}}}} &\raisebox{-0.45\height}{\includegraphics[width=0.19\textwidth,valign=m]{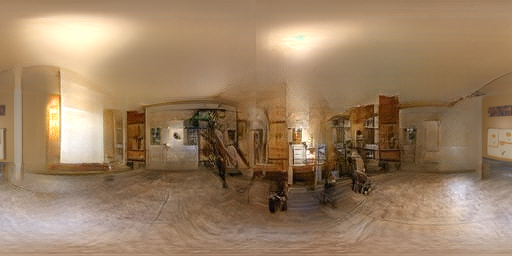}} &\raisebox{-0.45\height}{\includegraphics[width=0.19\textwidth,valign=m]{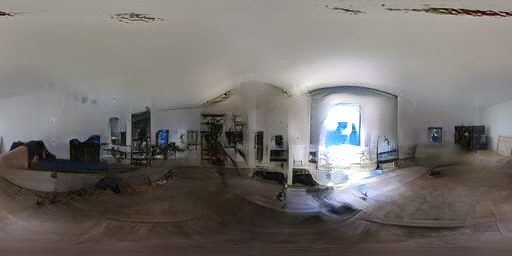}} &\raisebox{-0.45\height}{\includegraphics[width=0.19\textwidth,valign=m]{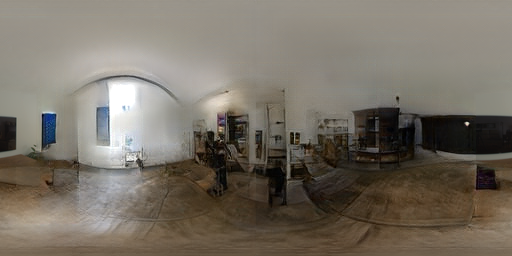}} &\raisebox{-0.45\height}{\includegraphics[width=0.19\textwidth,valign=m]{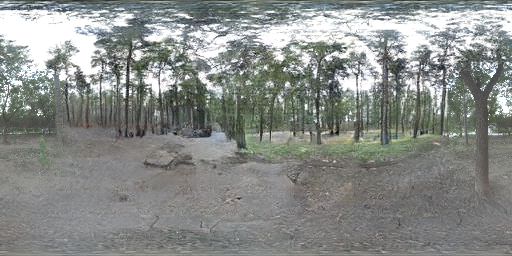}} &\raisebox{-0.45\height}{\includegraphics[width=0.19\textwidth,valign=m]{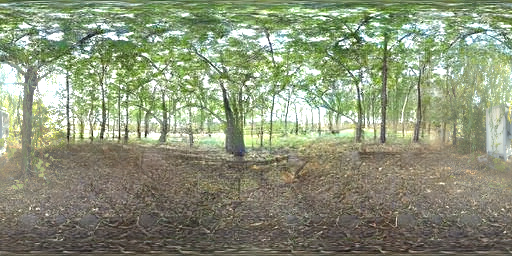}}\\

& \rule{0pt}{0.1pt} & \rule{0pt}{0.1pt} & \rule{0pt}{0.1pt} & \rule{0pt}{0.1pt} & \rule{0pt}{0.1pt} \\

&\includegraphics[width=0.19\textwidth,valign=m]{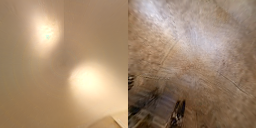} &\includegraphics[width=0.19\textwidth,valign=m]{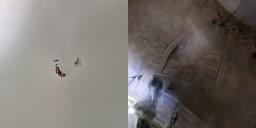} &\includegraphics[width=0.19\textwidth,valign=m]{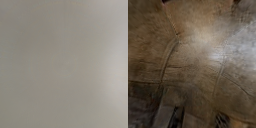} &\includegraphics[width=0.19\textwidth,valign=m]{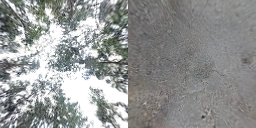} &\includegraphics[width=0.19\textwidth,valign=m]{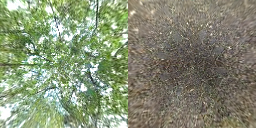}\\

& \rule{0pt}{0.1pt} & \rule{0pt}{0.1pt} & \rule{0pt}{0.1pt} & \rule{0pt}{0.1pt} & \rule{0pt}{0.1pt} \\

\multirow{3}{*}{\rotatebox{90}{\textbf{360U-Former\cite{Hilliard2024}}}}
&\raisebox{-0.7\height}{\includegraphics[width=0.19\textwidth,valign=m]{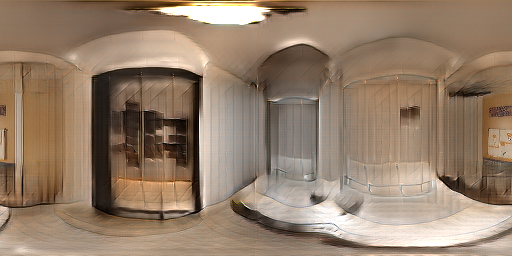}} &\raisebox{-0.7\height}{\includegraphics[width=0.19\textwidth,valign=m]{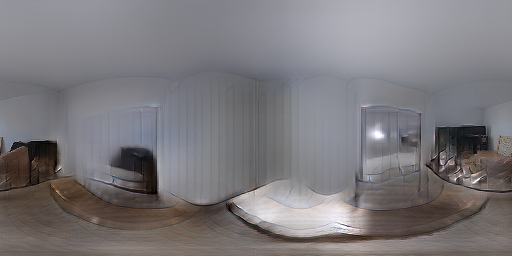}} &\raisebox{-0.7\height}{\includegraphics[width=0.19\textwidth,valign=m]{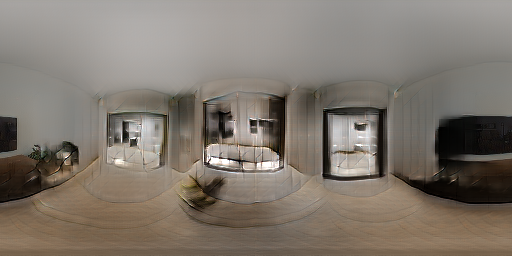}} &\raisebox{-0.7\height}{\includegraphics[width=0.19\textwidth,valign=m]{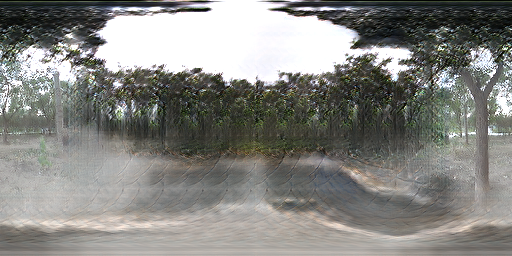}} &\raisebox{-0.7\height}{\includegraphics[width=0.19\textwidth,valign=m]{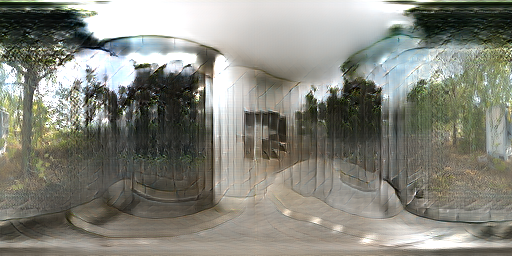}}\\

& \rule{0pt}{0.1pt} & \rule{0pt}{0.1pt} & \rule{0pt}{0.1pt} & \rule{0pt}{0.1pt} & \rule{0pt}{0.1pt} \\

&\includegraphics[width=0.19\textwidth,valign=m]{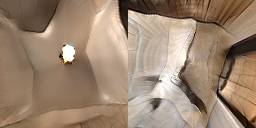} &\includegraphics[width=0.19\textwidth,valign=m]{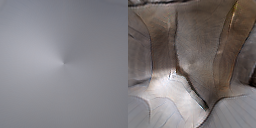} &\includegraphics[width=0.19\textwidth,valign=m]{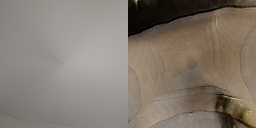} &\includegraphics[width=0.19\textwidth,valign=m]{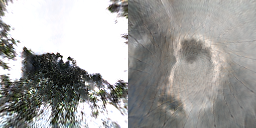} &\includegraphics[width=0.19\textwidth,valign=m]{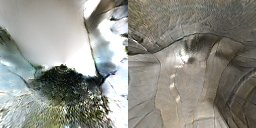}\\

& \rule{0pt}{0.1pt} & \rule{0pt}{0.1pt} & \rule{0pt}{0.1pt} & \rule{0pt}{0.1pt} & \rule{0pt}{0.1pt} \\

\multirow{3}{*}{\rotatebox{90}{\textbf{Ours (U-Net)}}} &\raisebox{-0.3\height}{\includegraphics[width=0.19\textwidth,valign=m]{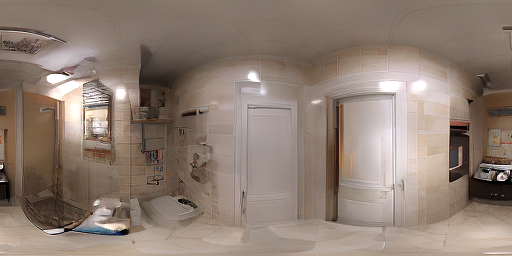}} &\raisebox{-0.3\height}{\includegraphics[width=0.19\textwidth,valign=m]{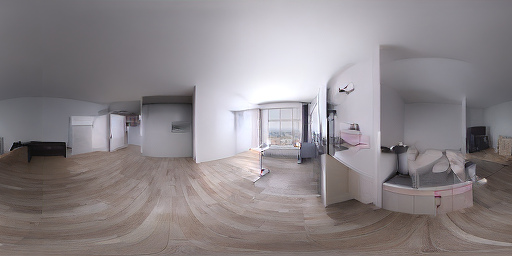}} &\raisebox{-0.3\height}{\includegraphics[width=0.19\textwidth,valign=m]{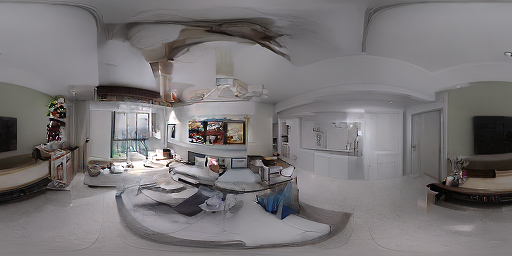}} &\raisebox{-0.3\height}{\includegraphics[width=0.19\textwidth,valign=m]{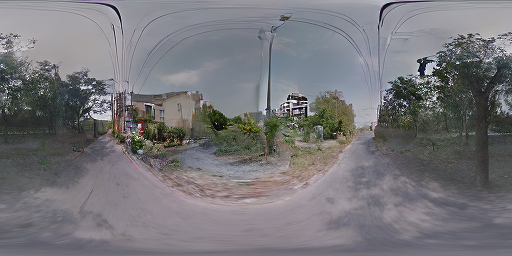}} &\raisebox{-0.3\height}{\includegraphics[width=0.19\textwidth,valign=m]{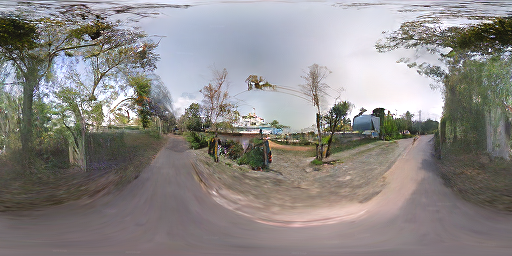}}\\

& \rule{0pt}{0.1pt} & \rule{0pt}{0.1pt} & \rule{0pt}{0.1pt} & \rule{0pt}{0.1pt} & \rule{0pt}{0.1pt} \\

&\includegraphics[width=0.19\textwidth,valign=m]{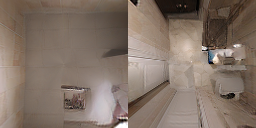} &\includegraphics[width=0.19\textwidth,valign=m]{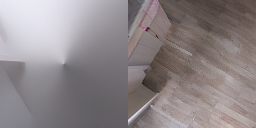} &\includegraphics[width=0.19\textwidth,valign=m]{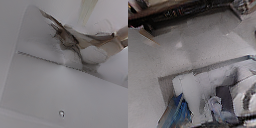} &\includegraphics[width=0.19\textwidth,valign=m]{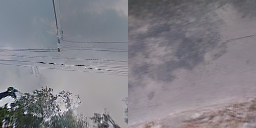} &\includegraphics[width=0.19\textwidth,valign=m]{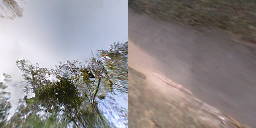}\\

& \rule{0pt}{0.1pt} & \rule{0pt}{0.1pt} & \rule{0pt}{0.1pt} & \rule{0pt}{0.1pt} & \rule{0pt}{0.1pt} \\

\multirow{3}{*}{\rotatebox{90}{\textbf{Ours (PanoDiT)}}} &\raisebox{-0.55\height}{\includegraphics[width=0.19\textwidth,valign=m]{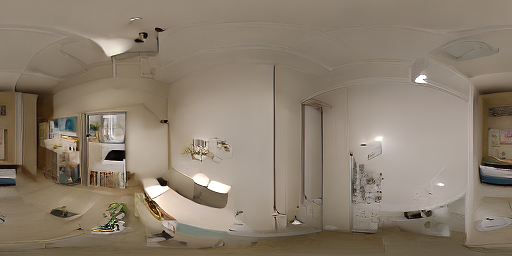}} &\raisebox{-0.55\height}{\includegraphics[width=0.19\textwidth,valign=m]{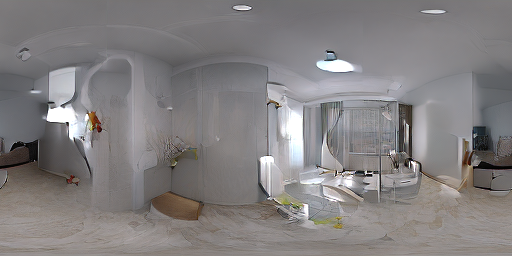}} &\raisebox{-0.55\height}{\includegraphics[width=0.19\textwidth,valign=m]{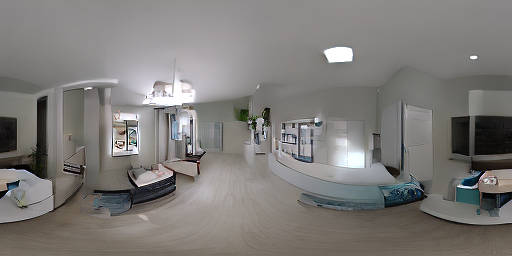}} &\raisebox{-0.55\height}{\includegraphics[width=0.19\textwidth,valign=m]{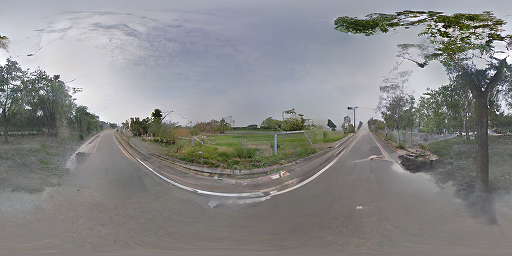}} &\raisebox{-0.55\height}{\includegraphics[width=0.19\textwidth,valign=m]{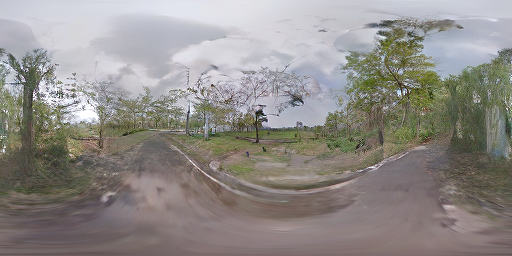}}\\

& \rule{0pt}{0.1pt} & \rule{0pt}{0.1pt} & \rule{0pt}{0.1pt} & \rule{0pt}{0.1pt} & \rule{0pt}{0.1pt} \\

&\includegraphics[width=0.19\textwidth,valign=m]{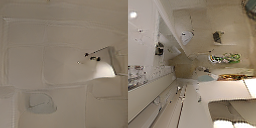} &\includegraphics[width=0.19\textwidth,valign=m]{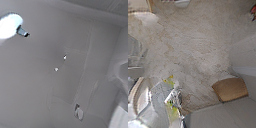} &\includegraphics[width=0.19\textwidth,valign=m]{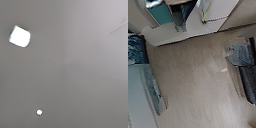} &\includegraphics[width=0.19\textwidth,valign=m]{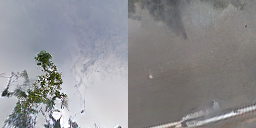} &\includegraphics[width=0.19\textwidth,valign=m]{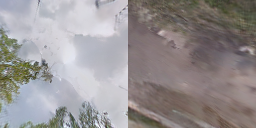}
\end{tabularx}
}
\vspace{-1em}
\caption{\small Indoor qualitative comparison of our method with state-of-the-art methods. For each method and input \acs{lfov} image, we show the \acs{erp} rotated 180$^\circ$, to show potential border seams, and the top and bottom cube map faces, to visualise generation at the poles.}
\label{fig:LDM_renders}
\end{figure*}

\begin{figure*}[t]
\footnotesize
\centering
\setlength\tabcolsep{0.6pt}
\renewcommand{\arraystretch}{0.1}
\resizebox{\linewidth}{!}{\begin{tabularx}{\linewidth}{m{0.025\linewidth}ccccc}

\rotatebox{90}{\textbf{Input}}  &\raisebox{-0.1\height}{\centering \includegraphics[width=0.1\textwidth,valign=m]{images/Denoise_Ablation/Crops/AG8A5582-5920abaaf0_02_crop.png}} &\raisebox{-0.1\height}{\centering \includegraphics[width=0.1\textwidth,valign=m]{images/Denoise_Ablation/Crops/9C4A9816-e2bf756838_01_crop.png}} &\raisebox{-0.1\height}{\centering \includegraphics[width=0.1\textwidth,valign=m]{images/Denoise_Ablation/Crops/AG8A9666-cc44ce23bc_07_crop.png}} &\raisebox{-0.1\height}{\centering \includegraphics[width=0.1\textwidth,valign=m]{images/Denoise_Ablation/Crops/00212_00.png}} &\raisebox{-0.1\height}{\centering \includegraphics[width=0.1\textwidth,valign=m]{images/Denoise_Ablation/Crops/00490_00.png}}\\

& \rule{0pt}{0.1pt} & \rule{0pt}{0.1pt} & \rule{0pt}{0.1pt} & \rule{0pt}{0.1pt} & \rule{0pt}{0.1pt} \\

\multirow{3}{*}{\rotatebox{90}{\textbf{Ground Truth}}} &\raisebox{-0.35\height}{\includegraphics[width=0.19\textwidth,valign=m]{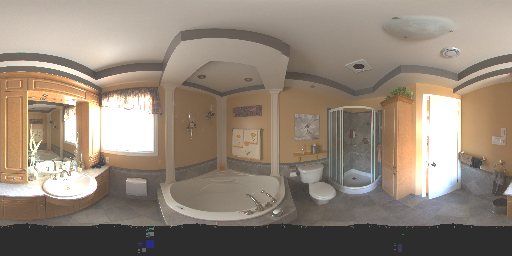}} &\raisebox{-0.35\height}{\includegraphics[width=0.19\textwidth,valign=m]{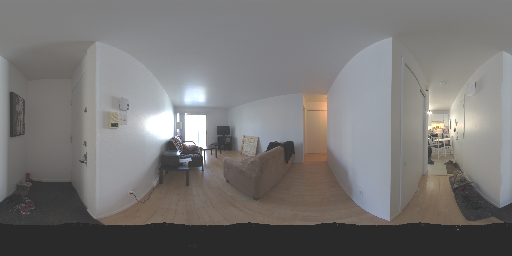}} &\raisebox{-0.35\height}{\includegraphics[width=0.19\textwidth,valign=m]{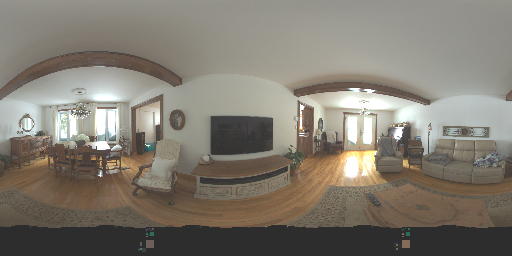}}&\raisebox{-0.35\height}{\includegraphics[width=0.19\textwidth,valign=m]{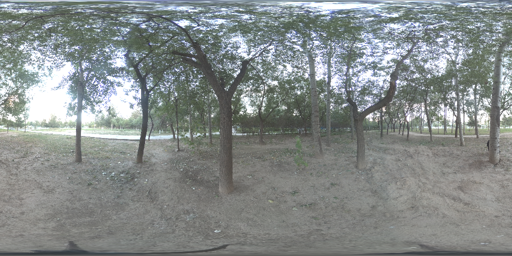}} &\raisebox{-0.35\height}{\includegraphics[width=0.19\textwidth,valign=m]{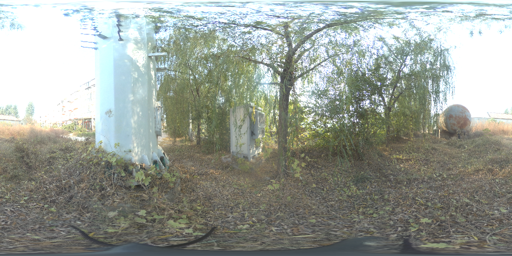}}\\

& \rule{0pt}{0.1pt} & \rule{0pt}{0.1pt} & \rule{0pt}{0.1pt} & \rule{0pt}{0.1pt} & \rule{0pt}{0.1pt} \\

&\includegraphics[width=0.19\textwidth,valign=m]{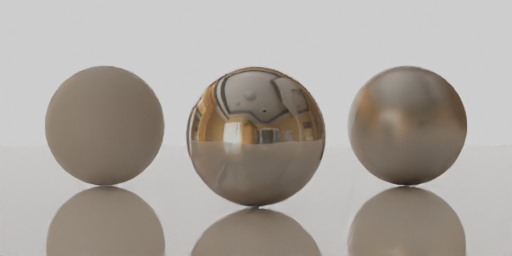} &\includegraphics[width=0.19\textwidth,valign=m]{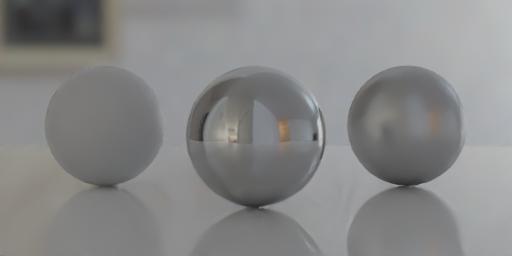} &\includegraphics[width=0.19\textwidth,valign=m]{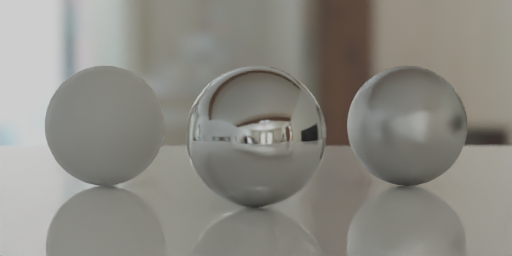}&\includegraphics[width=0.19\textwidth,valign=m]{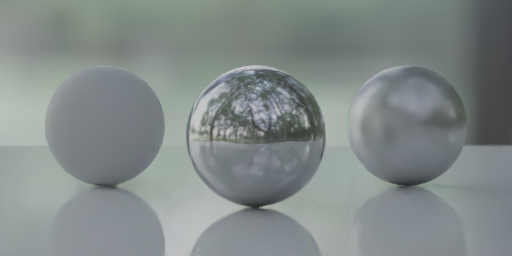} &\includegraphics[width=0.19\textwidth,valign=m]{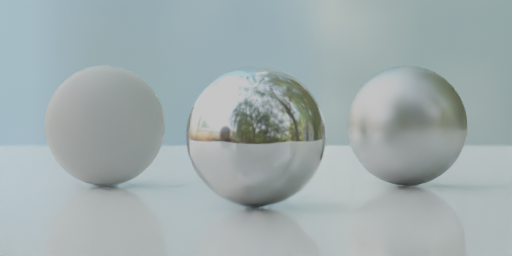}\\

& \rule{0pt}{0.1pt} & \rule{0pt}{0.1pt} & \rule{0pt}{0.1pt} & \rule{0pt}{0.1pt} & \rule{0pt}{0.1pt} \\

\multirow{3}{*}{\rotatebox{90}{\textbf{Weber '22 \cite{Weber2022}}}} &\raisebox{-0.5\height}{\includegraphics[width=0.19\textwidth,valign=m]{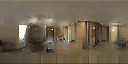}} &\raisebox{-0.5\height}{\includegraphics[width=0.19\textwidth,valign=m]{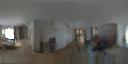}} &\raisebox{-0.5\height}{\includegraphics[width=0.19\textwidth,valign=m]{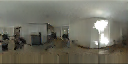}}\\

& \rule{0pt}{0.1pt} & \rule{0pt}{0.1pt} & \rule{0pt}{0.1pt} & \rule{0pt}{0.1pt} & \rule{0pt}{0.1pt} \\

&\includegraphics[width=0.19\textwidth,valign=m]{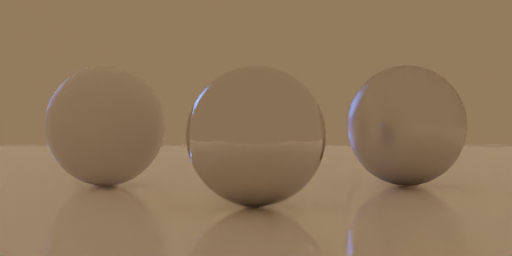} &\includegraphics[width=0.19\textwidth,valign=m]{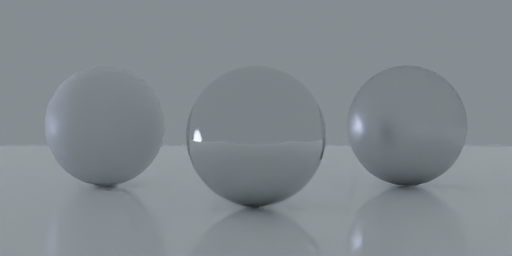} &\includegraphics[width=0.19\textwidth,valign=m]{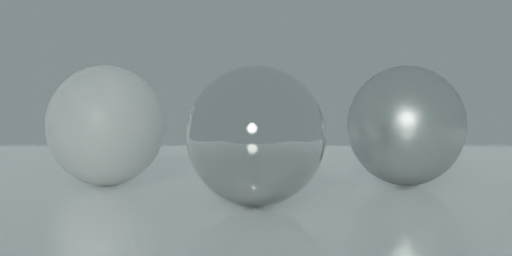}\\

& \rule{0pt}{0.1pt} & \rule{0pt}{0.1pt} & \rule{0pt}{0.1pt} & \rule{0pt}{0.1pt} & \rule{0pt}{0.1pt} \\

\multirow{3}{*}{\rotatebox{90}{\textbf{EverLight \cite{Dastjerdi2023}}}} &\raisebox{-0.45\height}{\includegraphics[width=0.19\textwidth,valign=m]{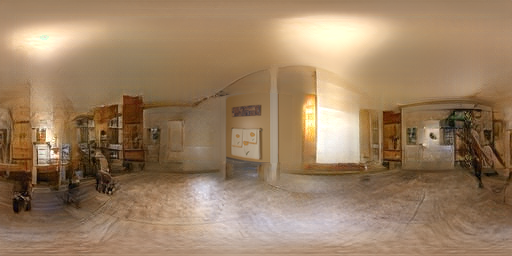}} &\raisebox{-0.45\height}{\includegraphics[width=0.19\textwidth,valign=m]{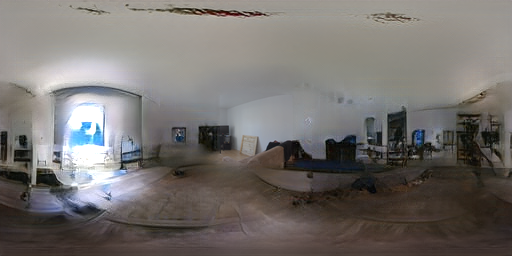}} &\raisebox{-0.45\height}{\includegraphics[width=0.19\textwidth,valign=m]{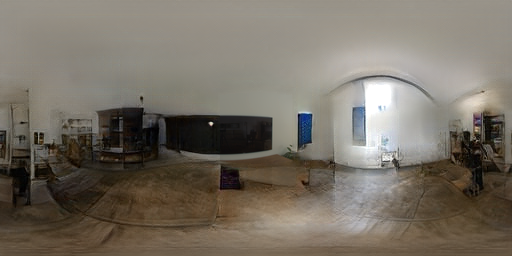}} &\raisebox{-0.45\height}{\includegraphics[width=0.19\textwidth,valign=m]{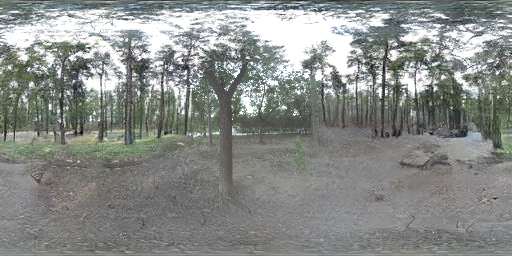}} &\raisebox{-0.45\height}{\includegraphics[width=0.19\textwidth,valign=m]{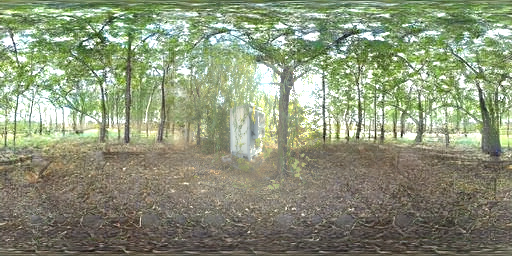}}\\

& \rule{0pt}{0.1pt} & \rule{0pt}{0.1pt} & \rule{0pt}{0.1pt} & \rule{0pt}{0.1pt} & \rule{0pt}{0.1pt} \\

&\includegraphics[width=0.19\textwidth,valign=m]{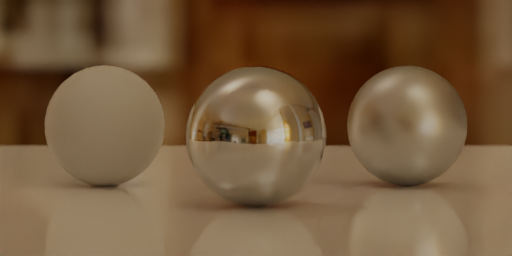} &\includegraphics[width=0.19\textwidth,valign=m]{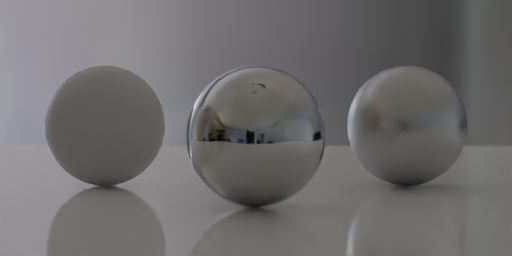} &\includegraphics[width=0.19\textwidth,valign=m]{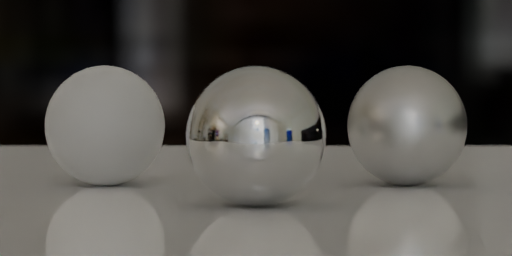} &\includegraphics[width=0.19\textwidth,valign=m]{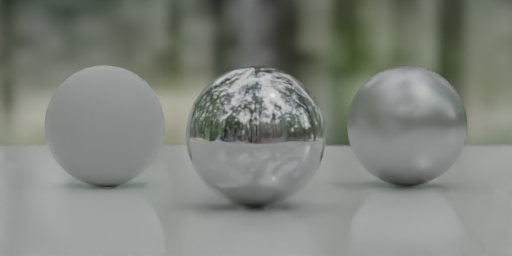} &\includegraphics[width=0.19\textwidth,valign=m]{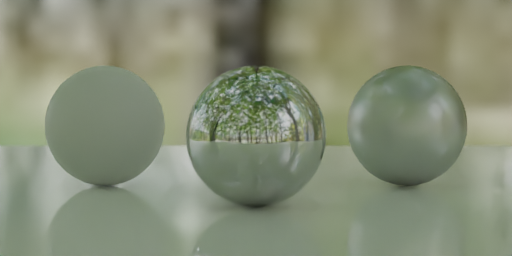}\\

& \rule{0pt}{0.1pt} & \rule{0pt}{0.1pt} & \rule{0pt}{0.1pt} & \rule{0pt}{0.1pt} & \rule{0pt}{0.1pt} \\

\multirow{3}{*}{\rotatebox{90}{\textbf{360U-Former\cite{Hilliard2024}}}}
&\raisebox{-0.7\height}{\includegraphics[width=0.19\textwidth,valign=m]{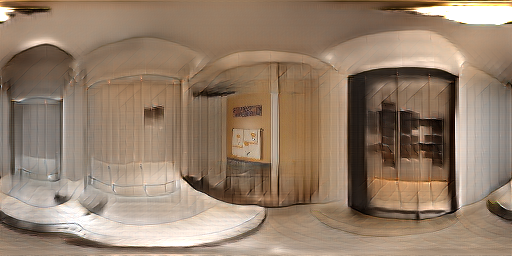}} &\raisebox{-0.7\height}{\includegraphics[width=0.19\textwidth,valign=m]{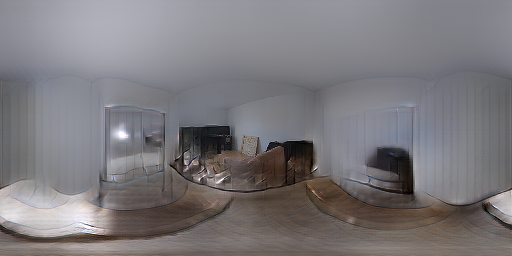}} &\raisebox{-0.7\height}{\includegraphics[width=0.19\textwidth,valign=m]{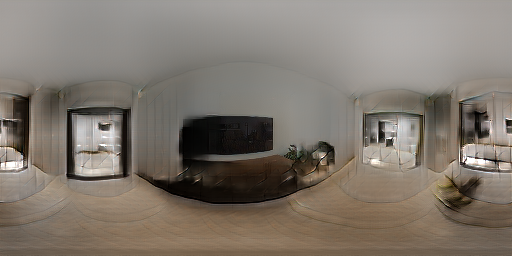}} &\raisebox{-0.7\height}{\includegraphics[width=0.19\textwidth,valign=m]{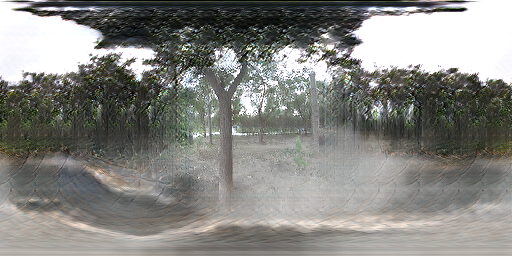}} &\raisebox{-0.7\height}{\includegraphics[width=0.19\textwidth,valign=m]{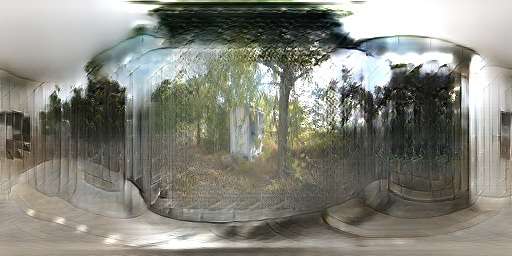}}\\

& \rule{0pt}{0.1pt} & \rule{0pt}{0.1pt} & \rule{0pt}{0.1pt} & \rule{0pt}{0.1pt} & \rule{0pt}{0.1pt} \\

&\includegraphics[width=0.19\textwidth,valign=m]{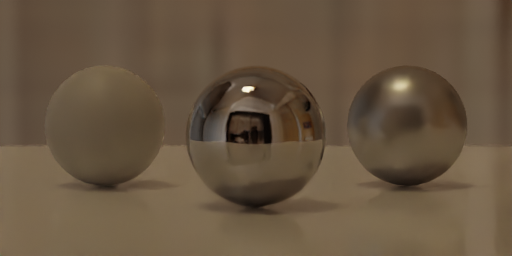} &\includegraphics[width=0.19\textwidth,valign=m]{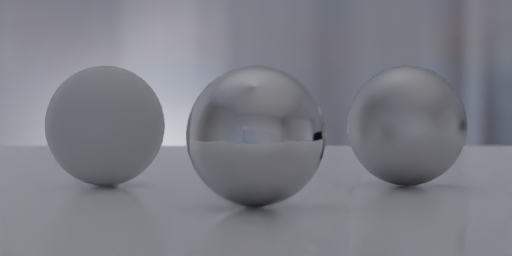} &\includegraphics[width=0.19\textwidth,valign=m]{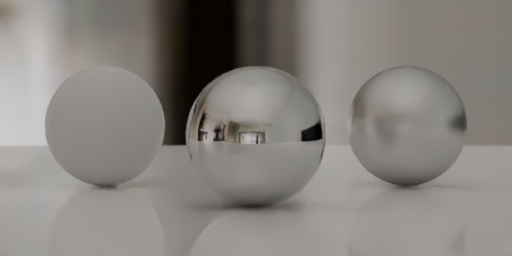} &\includegraphics[width=0.19\textwidth,valign=m]{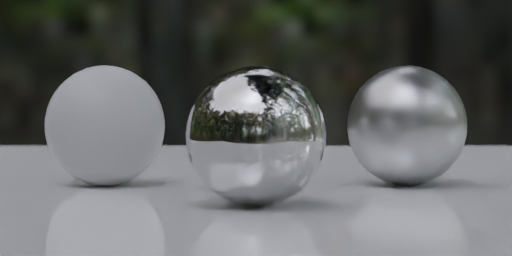} &\includegraphics[width=0.19\textwidth,valign=m]{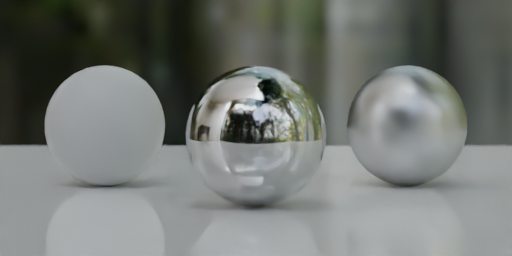}\\

& \rule{0pt}{0.1pt} & \rule{0pt}{0.1pt} & \rule{0pt}{0.1pt} & \rule{0pt}{0.1pt} & \rule{0pt}{0.1pt} \\

\multirow{3}{*}{\rotatebox{90}{\textbf{Ours (U-Net)}}} &\raisebox{-0.3\height}{\includegraphics[width=0.19\textwidth,valign=m]{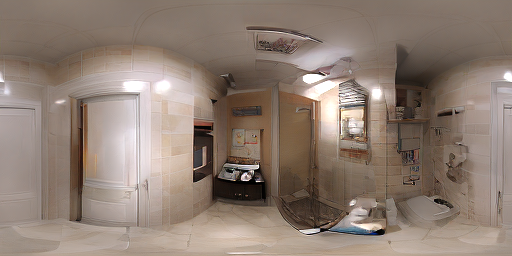}} &\raisebox{-0.3\height}{\includegraphics[width=0.19\textwidth,valign=m]{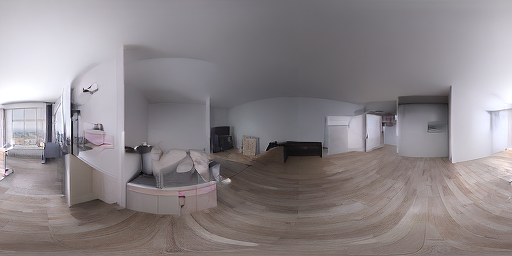}} &\raisebox{-0.3\height}{\includegraphics[width=0.19\textwidth,valign=m]{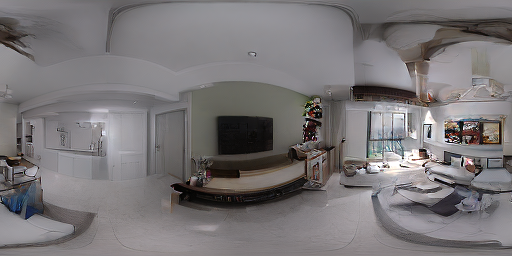}} &\raisebox{-0.3\height}{\includegraphics[width=0.19\textwidth,valign=m]{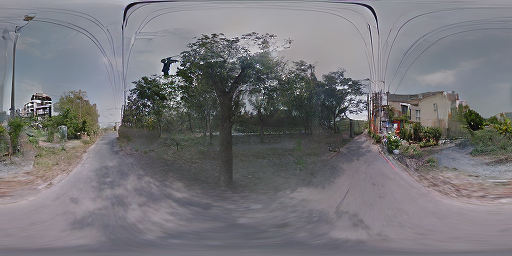}} &\raisebox{-0.3\height}{\includegraphics[width=0.19\textwidth,valign=m]{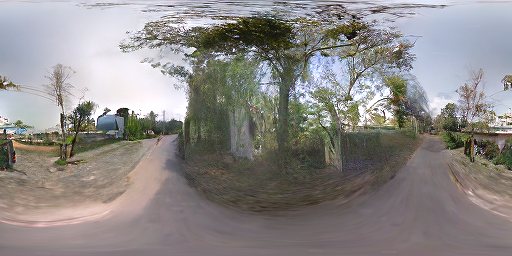}}\\

& \rule{0pt}{0.1pt} & \rule{0pt}{0.1pt} & \rule{0pt}{0.1pt} & \rule{0pt}{0.1pt} & \rule{0pt}{0.1pt} \\

&\includegraphics[width=0.19\textwidth,valign=m]{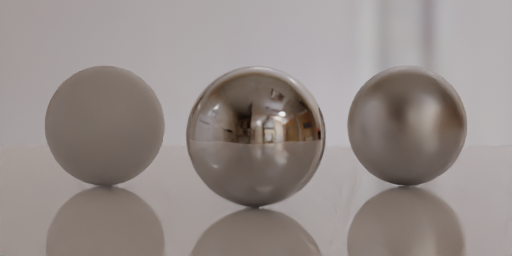} &\includegraphics[width=0.19\textwidth,valign=m]{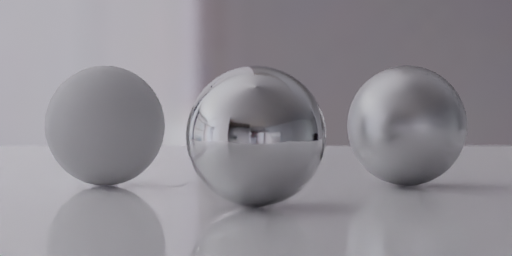} &\includegraphics[width=0.19\textwidth,valign=m]{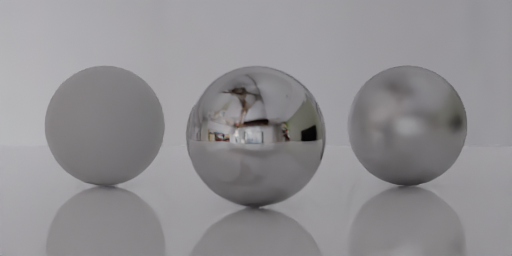} &\includegraphics[width=0.19\textwidth,valign=m]{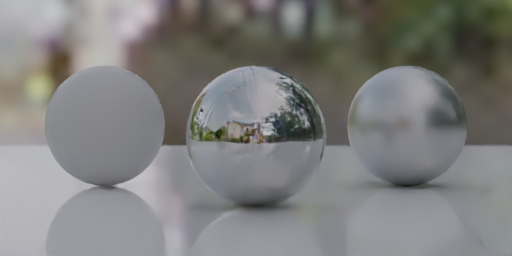} &\includegraphics[width=0.19\textwidth,valign=m]{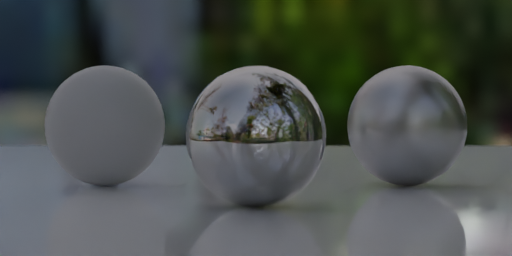}\\

& \rule{0pt}{0.1pt} & \rule{0pt}{0.1pt} & \rule{0pt}{0.1pt} & \rule{0pt}{0.1pt} & \rule{0pt}{0.1pt} \\

\multirow{3}{*}{\rotatebox{90}{\textbf{Ours (PanoDiT)}}} &\raisebox{-0.55\height}{\includegraphics[width=0.19\textwidth,valign=m]{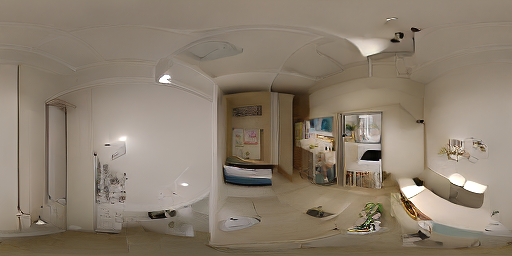}} &\raisebox{-0.55\height}{\includegraphics[width=0.19\textwidth,valign=m]{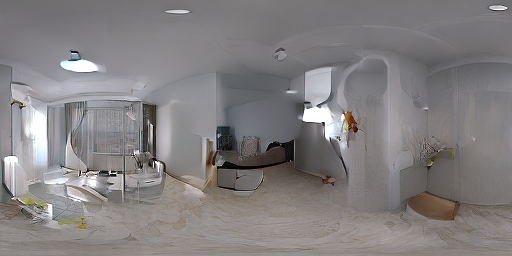}} &\raisebox{-0.55\height}{\includegraphics[width=0.19\textwidth,valign=m]{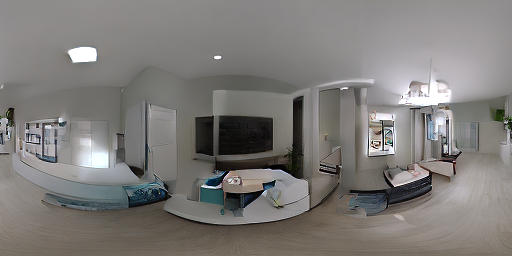}} &\raisebox{-0.55\height}{\includegraphics[width=0.19\textwidth,valign=m]{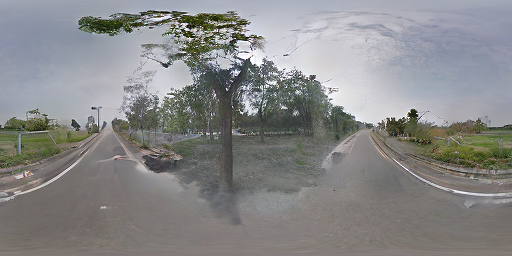}} &\raisebox{-0.55\height}{\includegraphics[width=0.19\textwidth,valign=m]{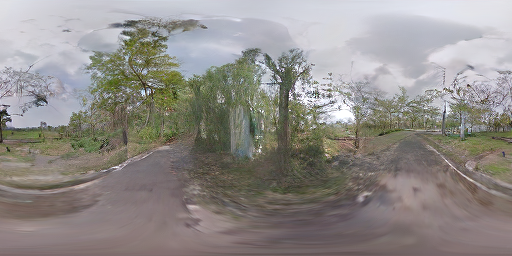}}\\

& \rule{0pt}{0.1pt} & \rule{0pt}{0.1pt} & \rule{0pt}{0.1pt} & \rule{0pt}{0.1pt} & \rule{0pt}{0.1pt} \\

&\includegraphics[width=0.19\textwidth,valign=m]{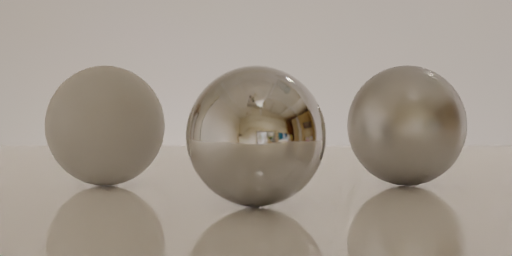} &\includegraphics[width=0.19\textwidth,valign=m]{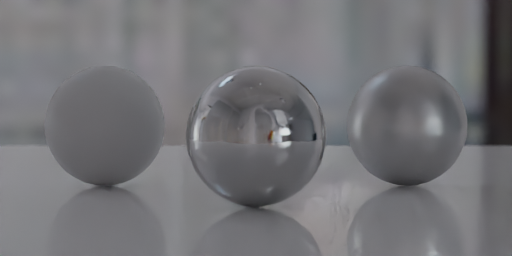} &\includegraphics[width=0.19\textwidth,valign=m]{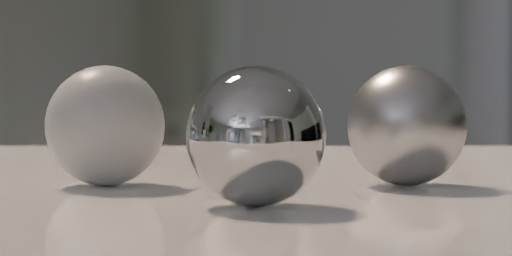} &\includegraphics[width=0.19\textwidth,valign=m]{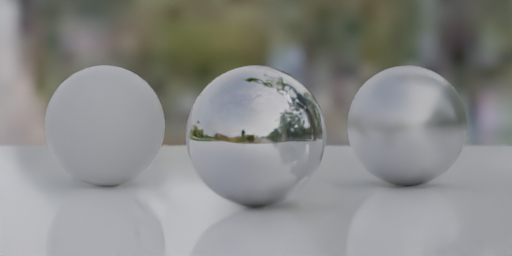} &\includegraphics[width=0.19\textwidth,valign=m]{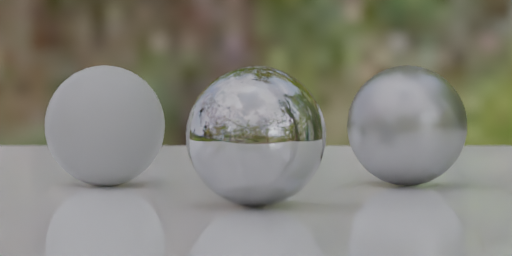}
\end{tabularx}
}
\vspace{-1em}
\caption{Indoor qualitative comparison of generated \acs{erp}s and spheres rendered with the \acs{erp}. The material of the spheres from \textit{Left to Right}: Diffuse, Mirror, Specular.}
\label{fig:LDM_spheres}
\end{figure*}

In \Cref{fig:LDM_renders}, we provide additional environment maps for both indoor and outdoor scenes from both of our models to compare with Weber'22~\cite{Weber2022}, EverLight~\cite{Dastjerdi2023} and 360U-Former~\cite{Hilliard2024}. Additionally, we render three spheres of different surface materials: diffuse, mirror-reflective and specular. This visualises how objects will appear when rendered by the environment maps and helps to assess estimated lighting accuracy against the state-of-the-art methods. The rendered spheres are presented alongside the generated environment maps in \Cref{fig:LDM_spheres}. When observing the rendered spheres of indoor scenes with our models, we observe colours similar to the ground-truth renders. However, in a few examples, light sources are out of place. For outdoor scenes, our models consistently and correctly estimate the sun position and the amount of cloud cover. These results reflect the quantitative results in \Cref{T_extra_LDM}.


\end{document}